\documentclass{article}
\usepackage{amssymb}

\usepackage[preprint]{corl_2025} 
\usepackage{amsmath} 
\usepackage{graphicx}
\usepackage{subcaption}
\usepackage{wrapfig} 
\usepackage{makecell}
\usepackage{siunitx}
\usepackage{booktabs}
\usepackage{algorithm}
\usepackage{algorithmic}
\newcommand{\fig}[1]{Fig.~\ref{#1}}
\newcommand{\tab}[1]{Table~\ref{#1}}
\newcommand{\eq}[1]{(\ref{#1})}

\title{Learning Pivoting Manipulation with Force and Vision Feedback Using Optimization-based Demonstrations}

\author{
  Yuki Shirai$^{1}$, Kei Ota$^{2}$, Devesh K. Jha$^{1}$, Diego Romeres$^{1}$\\
  $^{1}$Mitsubishi Electric Research Laboratories,  $^{2}$Mitsubishi Electric\\
}

\begin{document}
\maketitle


\begin{abstract}
Non-prehensile manipulation is challenging due to complex contact interactions between objects, the environment, and robots. 
Model-based approaches can efficiently generate complex trajectories of robots and objects under contact constraints. However, they tend to be sensitive to model inaccuracies and require access to privileged information (e.g., object mass, size, pose), making them less suitable for novel objects.
In contrast, learning-based approaches are typically more robust to modeling errors but require large amounts of data. 
In this paper, we bridge these two approaches to propose a framework for learning closed-loop pivoting manipulation. 
By leveraging computationally efficient Contact-Implicit Trajectory Optimization (CITO), we design demonstration-guided deep Reinforcement Learning (RL), leading to sample-efficient learning.
We also present a sim-to-real transfer approach using a privileged training strategy, enabling the robot to perform pivoting manipulation using only proprioception, vision, and force sensing without access to privileged information.
Our method is evaluated on several pivoting tasks, demonstrating that it can successfully perform sim-to-real transfer.
The overview of our method and the hardware experiments are shown \href{https://youtu.be/akjGDgfwLbM?si=QVw6ExoPy2VsU2g6}{here}.
\end{abstract}



\section{Introduction}

Non-prehensile manipulation, such as pivoting, pushing, and sliding, plays an important role in enhancing the dexterity of robotic systems \cite{billard2019trends, mason2018toward, rodriguez2021unstable}.
These skills allow robots to interact with the environment more flexibly, enabling them to adapt to a wide range of tasks without requiring secure grasps.
However, achieving such skills is challenging due to the inherently complex contact interactions (e.g., making-breaking contact, sliding-sticking contact).
These interactions introduce non-smooth dynamics that are difficult to model and control as the number of contacts increases.

Model-based optimization methods, such as CITO and Model Predictive Control (MPC) \cite{sleiman2019contact, pang2023global, le2024fast, moura2022non, wijayarathne2023real, 9113247}, have demonstrated impressive performance, particularly in generating diverse trajectories at low computational cost. However, since these methods, in general, rely on simplified models of manipulation, they can be highly sensitive to uncertainties due to model inaccuracies. More critically, they often rely on offline system identification or online estimation of privileged information, such as object properties or contact states. This dependency limits the applicability of model-based controllers, particularly in real-world scenarios involving novel objects or partially observable environments.

Learning-based methods, such as RL, have also shown impressive performance, especially in their robustness against various sources of uncertainty \cite{levine2016end, Rajeswaran-RSS-18, beltran2020variable, lee2020learning, andrychowicz2020learning, handa2023dextreme, qi2023general}. These methods can operate without privileged information by directly learning policies from raw observations.
However, they typically require a large number of training samples, resulting in long training times, which poses a significant challenge for practical deployment.
This is especially problematic in non-prehensile manipulation, where the policy must reason object pose, contact locations, contact forces, and feasible action spaces from indirect and partial observations. 
Unlike prehensile manipulation (e.g., grasping \cite{mason2018toward}), where grasping provides stable control, non-prehensile tasks often involve underactuated dynamics and complex contact constraints that make the learning problem significantly harder. As a result, RL may often fail to discover viable solutions within a reasonable training time.

\begin{figure}[t]
    \centering
    \includegraphics[width=1.0\textwidth]{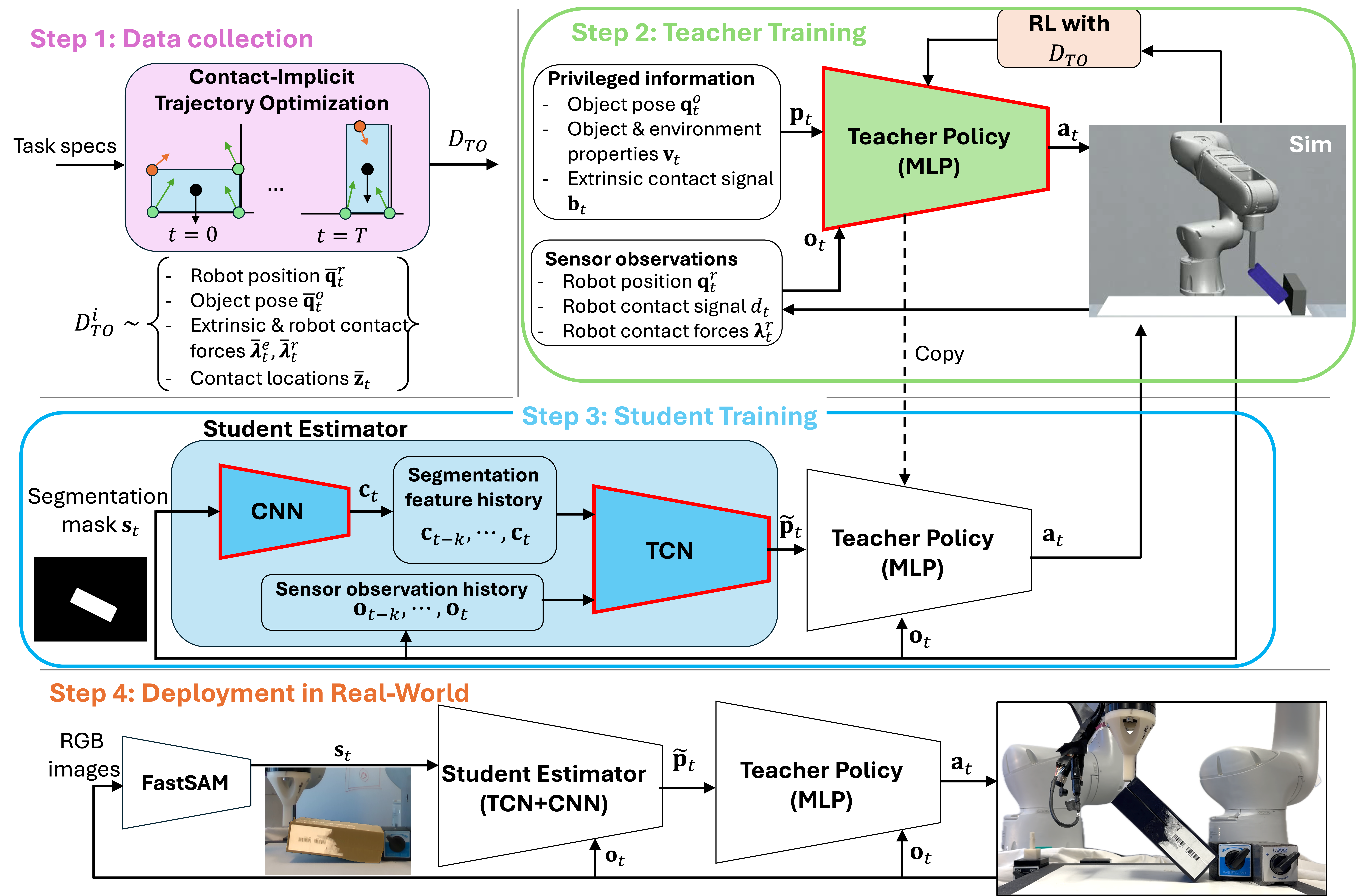}
    \caption{Overview of our proposed framework. Trainable modules have red edges. \textbf{Step 1}: We collect data using CITO given a user-specified task. \textbf{Step 2}: The teacher policy is trained using RL with privileged information and sensor observations, leveraging the demonstrations collected in Step 1. \textbf{Step 3}: The student estimator is trained to estimate the privileged information. The estimator consists of a CNN and a TCN to process temporal sensor observations, including segmentation and force measurements.
    \textbf{Step 4}: During deployment in real-world, the learned student estimator and teacher policy run in zero-shot sim-to-real transfer on physical hardware. 
    }
\label{fig:overview}
\end{figure}

In this paper, we propose a framework that integrates the strengths of model-based planning with learning-based policy execution for non-prehensile pivoting manipulation.
Our approach employs a student-teacher paradigm \cite{chen2020learning}, as illustrated in \fig{fig:overview}. 
First, we employ CITO to collect a large number of task demonstrations across a range of privileged information parameters.
Second, a teacher policy is trained in a high-fidelity simulator using RL, leveraging the demonstrations (e.g., robot, object, \& contact trajectories) generated by CITO. By utilizing these demonstrations, the teacher policy achieves significantly higher sample efficiency than standard RL methods.
Third, we train a student estimator to predict the privileged information required by the teacher policy. During training, the student estimator takes as input the history of sensor observations and segmentation features extracted from the vision pipeline, enabling it to infer the privileged information.
Finally, we evaluate the trained policy in both simulation and hardware experiments, achieving zero-shot sim-to-real transfer.
We verify that our framework substantially improves training efficiency compared to existing baselines. Moreover, we verify that
our framework 
outperforms an MPC baseline, which struggles due to inaccuracies in privileged information. 
Our contributions are as follows.
    \begin{itemize}
        \item We propose a framework for learning contact-rich non-prehensile manipulation controllers and estimators by leveraging demonstrations generated by CITO.
        \item We develop a sim-to-real transfer approach based on a student-teacher architecture, where the student estimates privileged information from partial observations using a temporal history of visual and force sensing. 
        \item We demonstrate that our method achieves robust manipulation performance against various uncertainties (e.g., object physical parameters) in real-world experiments. 
    \end{itemize}


\section{Related Work}
\textbf{Model-Based Optimization for Contact-Rich Manipulation}.
Model-based optimization methods have successfully achieved various non-prehensile manipulation skills, such as pushing \cite{moura2022non, pang2023global, aydinoglu2024consensus}, pivoting \cite{aceituno2020global, shirai2024robust, shirai2022robust}, and pulling \cite{hogan2020tactile, jin2021trajectory}. 
These methods design manipulation skills computationally efficiently by leveraging techniques such as contact smoothing \cite{pang2023global, shirai2024linear}, mixed-integer convex optimization \cite{aceituno2020global, hogan2020reactive}, and distributed optimization \cite{aydinoglu2024consensus, shirai_2022iros}.
%
However, these methods typically require privileged information (i.e., full-state feedback).  For example, \citet{aydinoglu2021stabilization} assumes that contact forces in extrinsic contacts between the object and the environment are directly measurable, which becomes increasingly impractical as task complexity grows. 
In this paper, we relax the full-state feedback assumptions by adopting an RL approach, while still leveraging CITO to generate a large number of demonstrations. This strategy enables the agent to learn manipulation skills significantly more efficiently than standard RL methods that rely solely on sparse rewards. 

\textbf{Learning-Based Methods for Contact-Rich Manipulation}. Learning-based methods, such as RL, Imitation Learning (IL), and foundation model-based methods, have demonstrated remarkable success in robotic manipulation \cite{gu2017deep, chen2023visual, chi2023diffusion, fu2024mobile, black2024pi_0, team2025gemini, xu2025unit, lin2025sim, noseworthy2025forge, seo2023deep}, enabling complex tasks such as bimanual cable manipulation and folding laundry. 
However, all of these methods require a large number of training samples, resulting in prohibitively long training times. 

To improve sample efficiency, demonstration-guided RL has been studied \cite{vecerik2017leveraging, Rajeswaran-RSS-18, ankile2024imitation}, where the demonstrations are used to guide exploration of RL agent to learn the policy and improve sample efficiency. 
For example, \citet{ota2021deep} uses Rapidly-exploring Random Trees (RRT) and \citet{xiong2021learning} uses human videos for generating kinematically feasible demonstrations for manipulation. 
%
However, these works \cite{ota2021deep, xiong2021learning, zhang2018deep} only consider kinematically feasible demonstrations. 
  Incorporating contact force information into demonstrations could be critical to learn fine manipulation due to very thin margins of error imposed by contact constraints.
Although some works have explored dynamically feasible demonstrations in locomotion tasks \cite{fuchioka2023opt, sleiman2024guided, pmlr-v270-bruedigam25a}, 
there has been relatively little work on applying such demonstrations to manipulation tasks.
This is due to the lack of a reliable module for generating dynamically feasible demonstrations considering extrinsic contact states in manipulation. 
Some works \cite{hou2024adaptive, chen2025dexforce} leverage human demonstrations to capture contact forces, but collecting such data at scale is challenging and often requires significant manual effort. In contrast, we use CITO to automatically generate robot, object, and contact force trajectories, providing richer supervision and greater scalability than human demonstrations.

Bridging the sim-to-real gap is another key challenge.
Privileged information used during training is often unavailable during real-world deployment.
Some prior works reconstruct privileged states using external sensors \cite{sleiman2024guided, pmlr-v270-bruedigam25a}, such as AprilTags \cite{5979561, ferrandis2024learning}.
The recent advances in the student-teacher framework \cite{chen2020learning, kumar2021rma, chen2023visual, miki2022learning, kaufmann2023champion, su2024sim2real, bauza2024demostart, jiang2024transic} enable zero-shot sim-to-real transfer by learning to predict privileged information.
Although some works have applied the student-teacher framework to manipulation, they often rely on restrictive assumptions---for example, assuming that object size remains constant \cite{fuchioka2024robotic, ferrandis2024learning}. 
In contrast, although we also adopt a student-teacher framework, we do not rely on such assumptions. By using a temporal history of force measurements and segmentation images, our student estimator is more broadly applicable to real-world scenarios involving novel objects.

\section{Method}\label{sec:method}
In this section, we present our proposed framework, as shown in \fig{fig:overview}. The objective is to learn pivoting manipulation using only proprioceptive, visual, and force sensing. The proposed framework consists of three steps. In Step 1, task demonstrations are generated using CITO. In Step 2, a teacher policy which has access to the privileged information is trained using RL with the sampled demonstrations collected in Step 1. 
In Step 3, a student estimator is trained to estimate the privileged information, which serves as input to the teacher policy. The teacher policy with the predictions of the trained student estimator is ultimately deployed on physical hardware for real-world validation. 

In this work, we make the following assumptions: (1) both the objects and the robots are rigid and the center of mass is located at the geometric centers, (2) manipulation occurs under quasi-static condition in SE(2), and (3) the robot end-effector pose, camera sensing, and robot contact force measurements are consistently available throughout manipulation.

\textbf{Step 1: Collecting Demonstrations Using Contact-Implicit Trajectory Optimization}
We collect a large set of datasets using CITO in \cite{shirai2025hierarchical}. For $N_r$ robots, we consider the following CITO:
\begin{equation}\label{trajopt}
\begin{aligned}
\min_{\bar{\mathbf{q}}_t,  \dot{\bar{\mathbf{q}}}_{t}, \bar{\mathbf{y}}_t} &  \sum_{t=0}^{T} \|\bar{\mathbf{q}}_{t} - \bar{\mathbf{q}}_t^\text{ref}\|^2_{Q}\\
\text{s. t. }, \, &  f_\text{dyn}\left(\bar{\mathbf{q}}_t, \bar{\mathbf{q}}_{t+1},  \dot{\bar{\mathbf{q}}}_{t}, \bar{\mathbf{y}}_t\right)  = \mathbf{0}, 
g\left(\bar{\mathbf{q}}_t, \dot{\bar{\mathbf{q}}}_{t}, \bar{\mathbf{y}}_t\right)  = \mathbf{0},  
\end{aligned}
\end{equation}
where $\bar{\mathbf{q}}_t:= [\bar{\mathbf{q}}_t^o, \bar{\mathbf{q}}_t^r]$ and $\bar{\mathbf{y}}_t:= [\bar{\boldsymbol{\lambda}}_t^e, \bar{\boldsymbol{\lambda}}_t^r, \bar{\mathbf{z}}_t]$. $\bar{\mathbf{q}}_t^o \in \mathbb{R}^3$ represent an object pose in SE(2) and $\bar{\mathbf{q}}_t^r \in \mathbb{R}^{2\times N_r}$ represent robot end-effector positions in SE(2), respectively. 
The end-effector orientation is kept fixed throughout the task.
$\bar{\boldsymbol{\lambda}}_t^e \in \mathbb{R}^{2\times N_e}$ and $\bar{\boldsymbol{\lambda}}_t^r \in \mathbb{R}^{2\times N_r}$ represent contact forces between an object and the environment, and between an object and the robots, respectively. $N_e$ represents the potential extrinsic number of contacts between the object and the environment. 
We denote $\bar{\mathbf{z}}_t \in \mathbb{R}^{2\times N_e}$ as the extrinsic contact location between the object and the environment. 
$\bar{\mathbf{q}}_t^\text{ref} \in \mathbb{R}^3$ represents the linear interpolation between the start and goal object pose with $T$ steps.
We use the subscript $t$ to represent the timestep $t$.
We denote $f_\text{dyn}$ as non-smooth dynamics of the non-prehensile manipulation, including nonsmooth contact switching, force and moment balance, and friction cone constraints.
We denote $g$ as non-dynamics related constraints, such as bounds of decision variables and collision-avoidance. 
We emphasize that the generation of trajectories that satisfy kinematic feasibility alone and not dynamic feasibility are simple to obtain by removing some of the $f_\text{dyn}$ constraints, such as force and moment balance constraints.
Thus, we denote kinematically feasible dynamics as $f_\text{kin}$.
The problem \eq{trajopt} is solved using solvers such as Gurobi \cite{gurobi} and SNOPT \cite{gill2005snopt}. 
See \cite{shirai2025hierarchical} and the appendix for more details. 
Solving \eq{trajopt} generates $N$ demonstrations $D_\text{TO}:=\{D_\text{TO}^i\}_{i=1}^N$, where $D_\text{TO}^i:=\{\{\bar{\mathbf{q}}_t\}_{t=0}^T,  \{\bar{\mathbf{y}}_t\}_{t=0}^T\}^i$.
While previous works (e.g., \cite{ota2021deep, xiong2021learning, zhang2018deep}) only consider $\bar{\mathbf{q}}_t$ with $f_\text{kin}$, this work explicitly considers $\bar{\mathbf{q}}_t$ and $\bar{\mathbf{y}}_t$ with $f_\text{dyn}$. In particular, $\bar{\boldsymbol{\lambda}}_t^r$ guides for agents to learn robot motion direction, while $\bar{\boldsymbol{\lambda}}_t^e$ and $\bar{\mathbf{z}}_t$ offer insights into preferred extrinsic contacts.

\textbf{Step 2: Learning Privileged Teacher-Policy}
In this step, a teacher policy is trained to achieve the desired pivoting manipulation in a simulation where privileged information is accessible. 
We formulate the problem as a Markov Decision Process (MDP), with each component defined as follows. 

\textbf{States.} 
States consist of the privileged and non-privileged information. 
The privileged information $\mathbf{p}_t$ includes the object pose ${\mathbf{q}}_t^o \in \mathbb{R}^3$, the object and environment properties $\mathbf{v}_t \in \mathbb{R}^{N_p}$, and the extrinsic contact signal $\mathbf{b}_t \in \mathbb{Z}^{N_e}$. 
The object pose ${\mathbf{q}}_t^o$ lies in the SE(2) and consists of two positional components and one orientation. 
 $\mathbf{v}_t$ encodes physical properties, which are the mass and size of the object, and the friction constants of both the object and the surrounding environment geometry.
The extrinsic contact signal $\mathbf{b}_t$ is a binary vector where each element indicates whether a specific face of the object is in contact with a predefined environment surface (e.g., wall, table).

The non-privileged information $\mathbf{o}_t$ consists of 
  the robot positions ${\mathbf{q}}_t^r \in \mathbb{R}^{2\times {N_r}}$, the binary robot contact signal ${d}_t \in \mathbb{Z}^{1}$, and the 2D contact forces ${\boldsymbol{\lambda}}_t^r \in \mathbb{R}^{2\times N_r}$ measured by force sensors mounted on the robots' wrists. 
All observations are approximately normalized to lie in the range $[-1, 1]$. 

\textbf{Actions.} We consider linear translational actions in SE(2) for each robot, denoted as $\mathbf{a}_t^{2\times {N_r}}$. Specifically, each action represents a relative position command for the robots' end-effector. These action commands are converted into joint torques using Operational Space Control (OSC) \cite{1087068}. 

\textbf{Rewards.} Based on how demonstrations are used, we consider three distinct reward formulations. 
We denote three different RL polices using different demonstrations (i.e., using different reward formulation) as (1) \emph{Vanilla RL}, which does not use any demonstrations, (2) \emph{Kinematics-conditioned RL}, and (3) \emph{Dynamics-conditioned RL}. These policies are obtained by 3 different rewards defined~as:
\begin{equation}
\begin{aligned}
\text{(1) Vanilla RL:} \quad & r = r_p + r_s + r_a \\
\text{(2) Kinematics-conditioned RL:} \quad & r = r_p + r_s + r_a + r_{\text{kin}} \\
\text{(3) Dynamics-conditioned RL:} \quad & r = r_p + r_s + r_a + r_{\text{dyn}}
\end{aligned}
\label{reward_eq}
\end{equation}
First, the progress reward is $r_p = \alpha_1\left(\frac{\pi}{2} -\theta_e\right) +  \alpha_2\left(\theta_e^2\right) $, where $\theta_e = \arccos{\left(\frac{1}{2}\left(Tr\left(R^\text{G}R\right)-1\right)\right)}$. 
$Tr(\cdot)$ denotes the matrix trace, and $R$ and $R^\text{G}$ are the goal and current rotation matrices, respectively. $\theta_e$ measures the angular deviation between the current and goal orientations, and $\frac{\pi}{2}$ is added as the offset. 
While the linear term in $r_p$ is used in \cite{zhang2023learning, zhang2023efficient}, our experiments reveal that the inclusion of the quadratic term is necessary to achieve higher success rates under domain randomization (DR)~\cite{tobin2017domain} over the size of the objects, which was not discussed in \cite{zhang2023efficient}.
Second, the sparse success reward is defined as $r_s = \alpha_3 \mathbb{I}_G\left(\mathbf{q}_t^o\right)$, where $\mathbb{I}_G$ is the indicator function over the goal set $G := \left\{ \mathbf{q}_t^o \in \mathbb{R}^3 \mid \|\mathbf{q}_t^o - \mathbf{q}_\text{goal}^o\| \leq \epsilon_s \right\}$, where $\mathbf{q}_\text{goal}^o$ is the desired goal state of the object and $\epsilon_s$ is the user-specified positive constant. 
Third, the action smoothness reward is given by $r_a = \alpha_4\|{\mathbf{a}}_{t-1} - {\mathbf{a}}_t\|^2$, for avoiding non-smooth actions.

Next, we define the reward based on demonstrations generated by CITO. For the kinematic reward $r_\text{kin}$, we use object and robot poses $\bar{\mathbf{q}}_t$ and extrinsic contact locations $\bar{\mathbf{z}}_t$ obtained by solving \eq{trajopt} with $f_\text{kin}$. Note that contact force demonstrations are not available in this setting, as $f_\text{kin}$ does not have dynamics constraints. Thus, we compute $r_{\text{kin}}$ as:
\begin{equation}
    r_{\text{kin}} = \alpha_5 ||{\mathbf{q}}_t^r - \phi(\mathbf{q}_t^o)||^2
\end{equation}
where $\phi$ retrieves the closest reference robot configuration $\bar{\mathbf{q}}_t^r$ corresponding to the current object observation $\mathbf{q}_t^o$. 
Since both the object and environment parameters are sampled from a known dataset $D_\text{TO}$ during simulation, the corresponding object reference trajectory $\bar{\mathbf{q}}_t^o$ is known. 
Using the current observation, we identify the closest object configuration within this trajectory, and consequently retrieve the closest robot configuration. 
This reward term encourages the robot to follow the kinematically feasible behaviors. 

Similarly, we define the dynamics reward $r_{\text{dyn}}$ by utilizing the demonstration $\bar{\mathbf{q}}_t$ and $\bar{\mathbf{y}}_t$ obtained by solving \eq{trajopt} with the dynamics model $f_\text{dyn}$:
\begin{equation}
    r_{\text{dyn}} = \alpha_6 ||{\mathbf{q}}_t^r - \phi(\mathbf{q}_t^o)||^2 + \alpha_7 \arccos{\left(\frac{{\boldsymbol{\lambda}}_t^r \cdot \psi(\mathbf{q}_t^o) }{||\boldsymbol{\lambda}_t^r|||| \psi(\mathbf{q}_t^o)||}\right)} + \alpha_8\mathbf{b}_t
\end{equation}
where $\psi$ retrieves the closest reference robot contact forces $\bar{\boldsymbol{\lambda}}_t^r$ corresponding to $\mathbf{q}_t^o$, following the same logic as $\phi$.
This reward encourages the robot to follow the dynamically feasible behaviors. 
In particular, the arccosine term in $r_{\text{dyn}}$ encourages the robot to perform a similar contact force direction as the demonstration shows. Importantly, we do not enforce matching the magnitude of the contact force, as we observe significant discrepancies between the dynamics model by $f_\text{dyn}$ and those in simulators (e.g., MuJoCo), leading to a potential sim2sim gap in contact force magnitudes.
Hence, this work focuses on the direction of contact forces. 
The term $\mathbf{b}_t$ is used to count if the desired extrinsic contact states occur.
The constants $\alpha_{i=1, 2, 3, 8}$ are positive and the others are negative. 





\textbf{Step 3: Learning Student-Estimator}
The objective of this step is to train the student estimator only using sensor observations to predict the privileged information as shown in \fig{fig:overview}. 
We empirically observe that sensor observations alone are sufficient for the object whose geometry is in-distribution with the dataset. However, their reliability declines when there is uncertainty in object size, which is quite common when manipulating novel objects. To address this, we additionally incorporate vision inputs to improve the estimation of the privileged information. Directly using RGB images is avoided due to potential noise, and employing 3D point clouds is excluded due to their significant computational cost (see \cite{chen2023visual}).
Instead, we leverage the object segmentation $\mathbf{s}_t$ derived from the RGB image, providing a compact but informative representation of the object. 

Therefore, we define a student encoder that takes the history of the sensor observations, $[\mathbf{o}_{t-T}, \cdots, \mathbf{o}_{t}]$, and the history of the segmentation features, $[\mathbf{s}_{t-T}, \cdots, \mathbf{s}_{t}]$. Since $\mathbf{s}_t$ is high-dimensional, we first apply a Convolutional Neural Network (CNN) to compress the segmentation into a lower-dimensional feature representation  $\mathbf{c}_t$.
Using the temporal histories of $\mathbf{o}_{t}$ and $\mathbf{c}_{t}$, we use a Temporal Convolutional Network (TCN)~\cite{bai2018empirical} to estimate the privileged information. We train CNN and TCN jointly via supervised learning using datasets collected by rolling out the teacher policy in the simulator under domain randomization. The supervised learning objective is to minimize the following loss function:
\begin{equation}
    l= ||\mathbf{p}_t - \tilde{\mathbf{p}}_t||^2,
    \label{eq:loss_student}
\end{equation}
where $\mathbf{p}_t$ is the ground-truth privileged information and $\tilde{\mathbf{p}}_t$ is the estimated output from the student encoder.
It is worth noting that we do not initialize the history buffer with zeros at the beginning of the episode as other works do (e.g., \cite{fuchioka2024robotic, qi2023hand, zakka2025mujoco}). Instead, we populate the buffer by repeating the initial observation and include this initialization scheme in the supervised learning dataset, which was critical for the student estimator to achieve accurate performance.




	

\section{Experiment Setup}
\label{sec:Setup}

We validate our framework across two distinct tasks (see \fig{hardware_fig}): \textbf{Pivoting with Wall}: pivoting a box using an external wall, \textbf{Pivoting without Wall}: pivoting a box without relying on external support. 
For the latter task, the table surface must provide very high friction. In simulation, increasing the friction coefficient alone was insufficient to replicate the real-world behavior. As a workaround, we add a thin virtual wall of height \SI{1}{\milli\meter} to simulate the effect of high-friction contact (see \fig{fig:hardware_fig_taskB}).
We define a trial as successful if the final orientation error satisfies $|\theta_e| \leq 0.087$ rad (i.e., \SI{5}{\degree}).
We describe the setup for each module below, with additional details provided in the appendix. 

\textbf{Demonstration Setup.} We use the method proposed in \cite{shirai2025hierarchical}, randomizing object and environment parameters to generate diverse demonstrations. 
For all tasks, we randomize the mass of the object, the friction constant of the object and the environment, and the size of the object.
For each task, we collect $5000$ demonstrations, which can be computed within a few minutes using $30$ Intel i9-13900K CPU cores.

\textbf{Teacher Policy Setup.} We train the teacher policy in MuJoCo simulator \cite{todorov2012mujoco} using robosuite \cite{robosuite2020} as a wrapper. The agent is trained using Soft Actor Critic (SAC) \cite{haarnoja2018soft}, implemented with tf2rl \cite{ota2020tf2rl}. For SAC, we use Multi-Layer Perceptron (MLP) for both actor and critic networks. The simulation runs at 500 Hz, while the policy operates at 10 Hz. 
For each episode, we set the maximum episode length to $300$ steps.
Overall, training converges within 4 hours on a single NVIDIA RTX 4090. 
During training, we apply domain randomization over the objects' mass and size, the friction constants of both the object and the environment, and the controller gains used in OSC within robosuite. Furthermore, we introduce sensor noise to both privileged information and sensor observations to account for the estimation errors from the student estimator during deployment. 
%

\textbf{Student Estimator Setup.} 
We first rollout the trained teacher policy over $2000$ episodes and collect a dataset containing ground-truth privileged information, sensor observations, and corresponding segmentation images (640×480 resolution) of the object using MuJoCo's rendering functionality. 
During data collection, we augment the segmentation images by introducing noise, such as randomly flipping, translating, and rotating segmentation masks, to improve robustness. 
We then train the student estimator via behavior cloning, minimizing the loss function  \eq{eq:loss_student} over multiple epochs. We use $T=5$ step history of the observations for training corresponding to 0.5 second. 
Overall, training converges with 10 epochs (1 hour roughly), depending on the range of domain randomization. 


\textbf{Hardware Setup.}
We use a 6 DoF MELFA robot \cite{mitsubishielectricFactoryAutomation} equipped with a stiffness controller and a 6-axis force/torque sensor. This hardware enables users to get robot end-effector positions and the force measurements in the world frame. 
For object segmentation, we use FastSAM \cite{zhao2023fast} to generate multiple instance segmentations from an RGB image captured by an Intel RealSense D435 RGB-D camera \cite{intelrealsenseDepthCamera}. To identify the target object, we filter the segmented instances under their corresponding point cloud information, under the assumption that a rough estimate of the SE(2) plane is available, as we focus exclusively on SE(2) planar manipulation.


\textbf{Baselines.} We implement an MPC baseline that uses privileged information, including object mass, size, and friction (identified offline), and object pose (estimated via AprilTags). At each timestep, MPC solves \eq{trajopt} in a receding-horizon manner, running at the same frequency as the teacher policy.



\section{Results}\label{sec:result}

Throughout our experiments, we aim to address the following questions:
\begin{enumerate}
\item Do demonstrations generated by CITO facilitate more effective and efficient learning?
\item How does the teacher policy's performance vary with different demonstrations?
\item How robust is the teacher policy compared to a baseline model-based method?
\item How accurately can the student estimator predict the privileged information? 
\item Can the trained policies be successfully transferred to real-world hardware experiments?
\end{enumerate}


\begin{wrapfigure}{r}{0.65\textwidth}
    \centering
    \begin{subfigure}{0.48\linewidth}
        \centering
        \includegraphics[width=\linewidth]{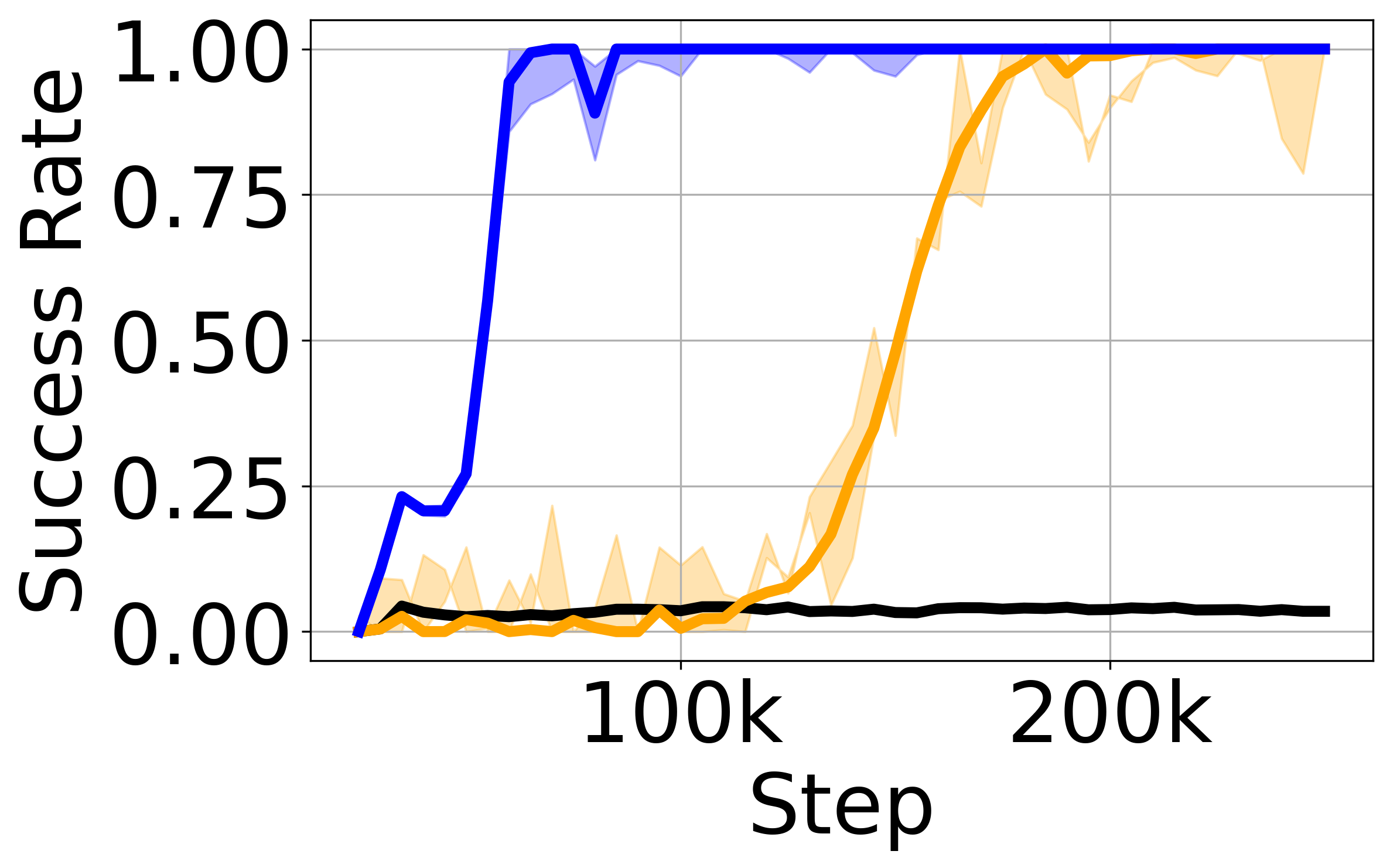}
        \caption{With external wall}
        \label{fig:with_wall}
    \end{subfigure}
    \hfill
    \begin{subfigure}{0.48\linewidth}
        \centering
        \includegraphics[width=\linewidth]{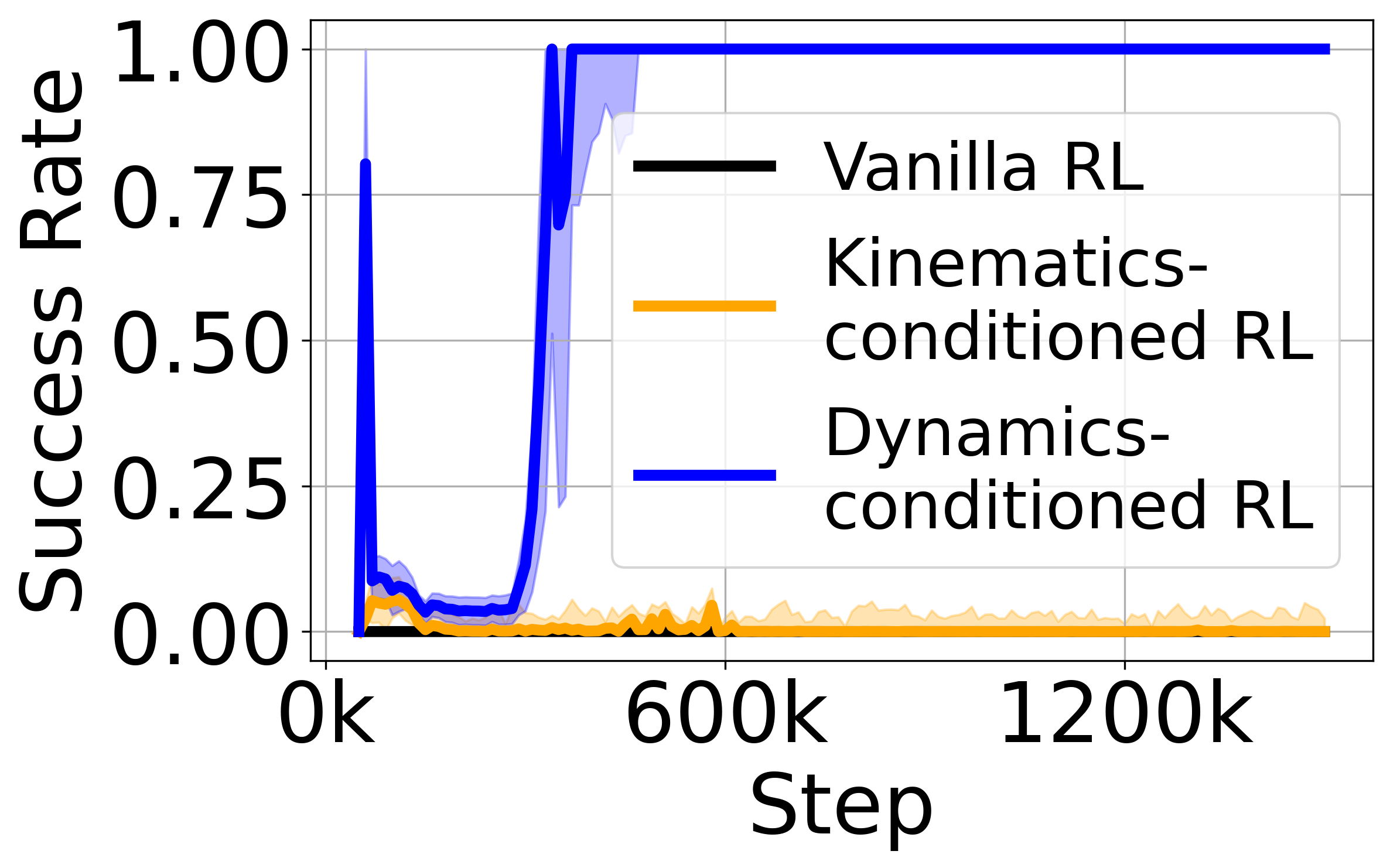}
        \caption{Without external wall}
        \label{fig:without_wall}
    \end{subfigure}
    \caption{Learning curves for different RL training runs. Solid lines indicate average success rates, and shaded regions denote standard deviation across three different random seeds. Every $10$k step, the current policy is evaluated over $50$ episodes, and the success rate is plotted.}
    \label{sample_eff}
\end{wrapfigure}

\textbf{Do demonstrations generated by CITO facilitate learning?}
Across the two tasks, we compare RL performance using different types of demonstrations, corresponding to the different reward formulations in \eq{reward_eq}.
RL with kinematic demonstrations is comparable to prior works such as \cite{ota2021deep, xiong2021learning}, which only consider kinematically feasible trajectories. 
Overall, RL with dynamics-based demonstrations achieves the fastest learning as shown in \fig{sample_eff}. 
In particular, in the pivoting without external wall task, neither vanilla RL nor kinematics-conditioned RL was able to learn the skill. 
We attribute this to the task's tighter feasible action space.
In contrast, dynamics-conditioned RL successfully learns the skill, benefiting from enriched demonstration with contact information.

\begin{wraptable}{r}{0.5\textwidth}
    \centering
    \caption{Number of successful attempts.}
    \begin{tabular}{rcc} \toprule
        Mass & \makecell{Kinematics-\\conditioned RL} & \makecell{Dynamics-\\conditioned RL} \\ \midrule
        $\SI{50}{\gram}$ & 2 / 3 &  \textbf{3} / 3  \\
         $\SI{110}{\gram}$ & \textbf{3} / 3 &  \textbf{3} / 3  \\
         $\SI{300}{\gram}$ & 0 / 3 &  \textbf{3} / 3  \\ \bottomrule
    \end{tabular}
    \label{tab:kin_RL-vs-dyn-RL}
\end{wraptable}

\textbf{How does the teacher policy's performance vary with different demonstrations?}
%
For the pivoting-with-wall task, we deploy both kinematics- and dynamics-conditioned RL policies on a real system using a box of mass \SI{110}{\gram}. During deployment, we vary the mass values used as privileged information. 
\tab{tab:kin_RL-vs-dyn-RL} shows the success rates over three trials.
We observe that dynamics-conditioned RL consistently outperforms kinematics-conditioned RL. 
While both policies are trained with access to privileged information, the dynamics-conditioned policy benefits from demonstrations that include contact force references. 
This enables the policy to learn physically grounded interaction behaviors during training, leading to greater robustness against variations in dynamic properties. 
In contrast, the kinematics-conditioned policy is trained with demonstrations that satisfy only geometric feasibility, making it more sensitive to changes in object properties.
These results highlight the importance of dynamics-aware demonstrations in contact-rich manipulation tasks.


\textbf{How robust is the learned policy compared to MPC?}
We compare the robustness of a dynamics-conditioned RL policy against an MPC controller on the real-world pivoting-with-wall task.
The true object length is \SI{0.16}{\meter}, and we introduce intentional mismatches in the assumed object length during deployment.
For example, a $-\SI{5}{\milli\meter}$ offset means that 
the actual size of the box is shorter than what the controllers expect. 
As shown in Table~\ref{tab:MPC-vs-RL}, both MPC and RL succeed when the actual object is longer than expected ($+\SI{5}{\milli\meter}$), as the contact with the wall is still maintained. However, when the actual object is shorter than expected ($-\SI{5}{\milli\meter}$), MPC fails completely, while RL remains successful. This suggests that the learned policy exhibits greater tolerance to moderate discrepancies in privileged information. At larger mismatches ($-\SI{10}{\milli\meter}$), even RL fails.
These results highlight the importance of accurate privileged information during deployment and motivate us to develop reliable estimators.
\begin{wraptable}{r}{0.4\textwidth}
    \centering
    \caption{Number of successful attempts.}
    \begin{tabular}{ccc} \toprule
          & \makecell{MPC} & \makecell{Dynamics-\\conditioned RL} \\ \midrule
        $+ \SI{5}{\milli\meter}$ & 3 / 3 &  3 / 3  \\
         $- \SI{5}{\milli\meter}$ & 0 / 3 &  3 / 3  \\
         $- \SI{10}{\milli\meter}$ & 0 / 3 &  0 / 3  \\ \bottomrule
    \end{tabular}
    \label{tab:MPC-vs-RL}
\end{wraptable}

\begin{figure}[t]
    \centering
    \begin{subfigure}{0.241\textwidth}
        \centering
        \includegraphics[width=1.0\linewidth]{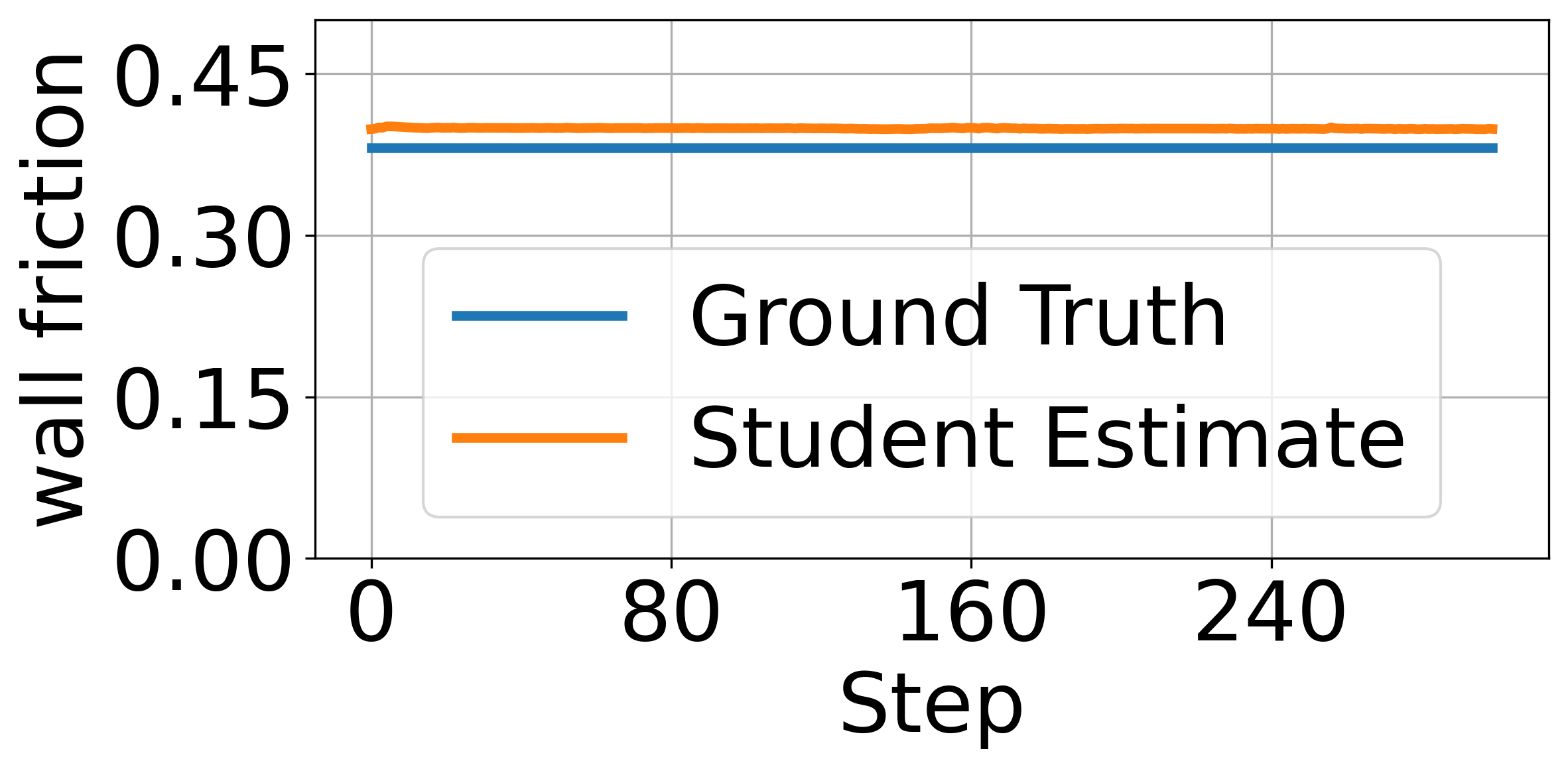}
    \end{subfigure}
    \begin{subfigure}{0.241\textwidth}
        \centering
        \includegraphics[width=1.0\linewidth]{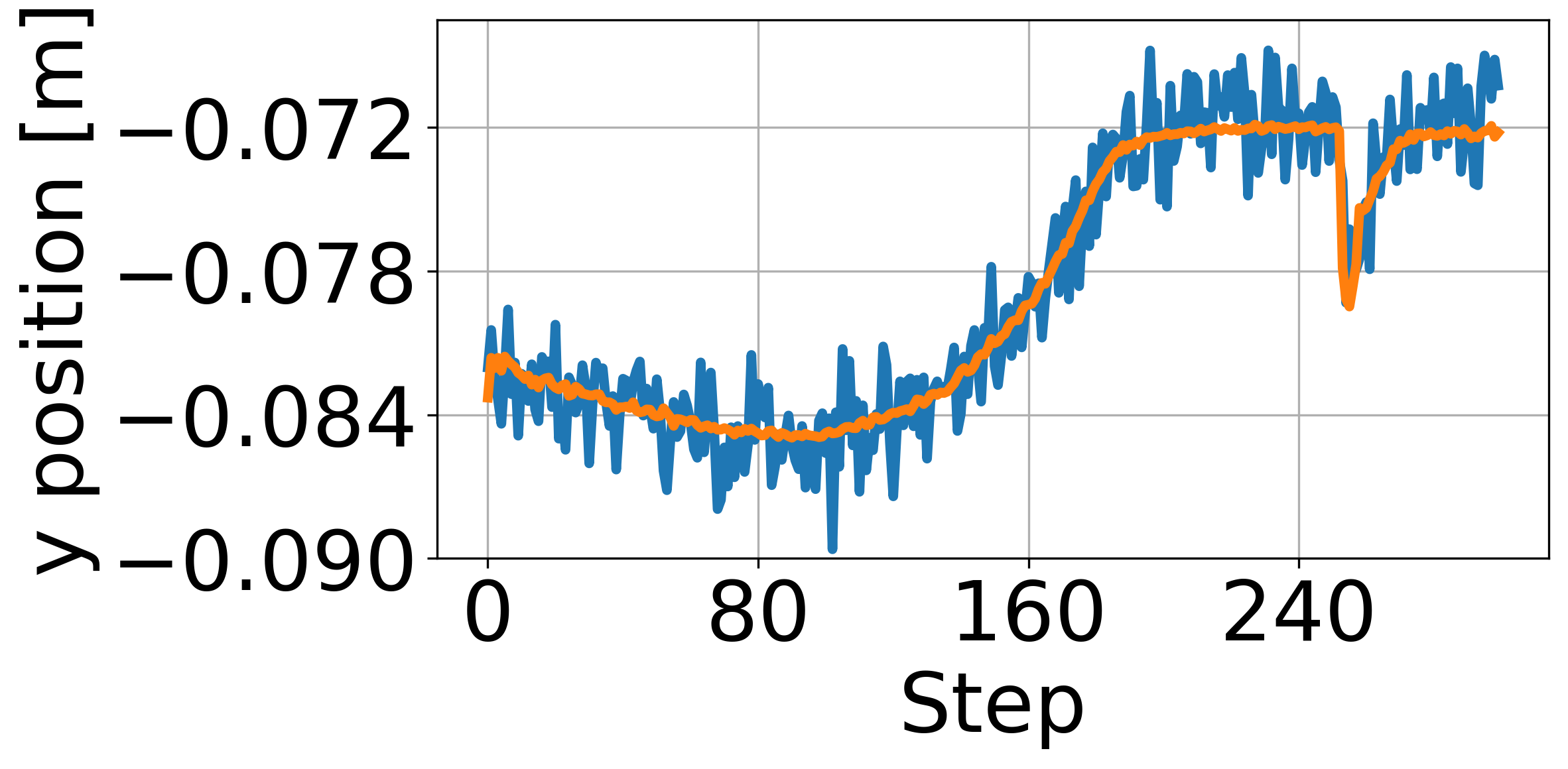}
    \end{subfigure}
        \begin{subfigure}{0.241\textwidth}
        \centering
        \includegraphics[width=1.0\linewidth]{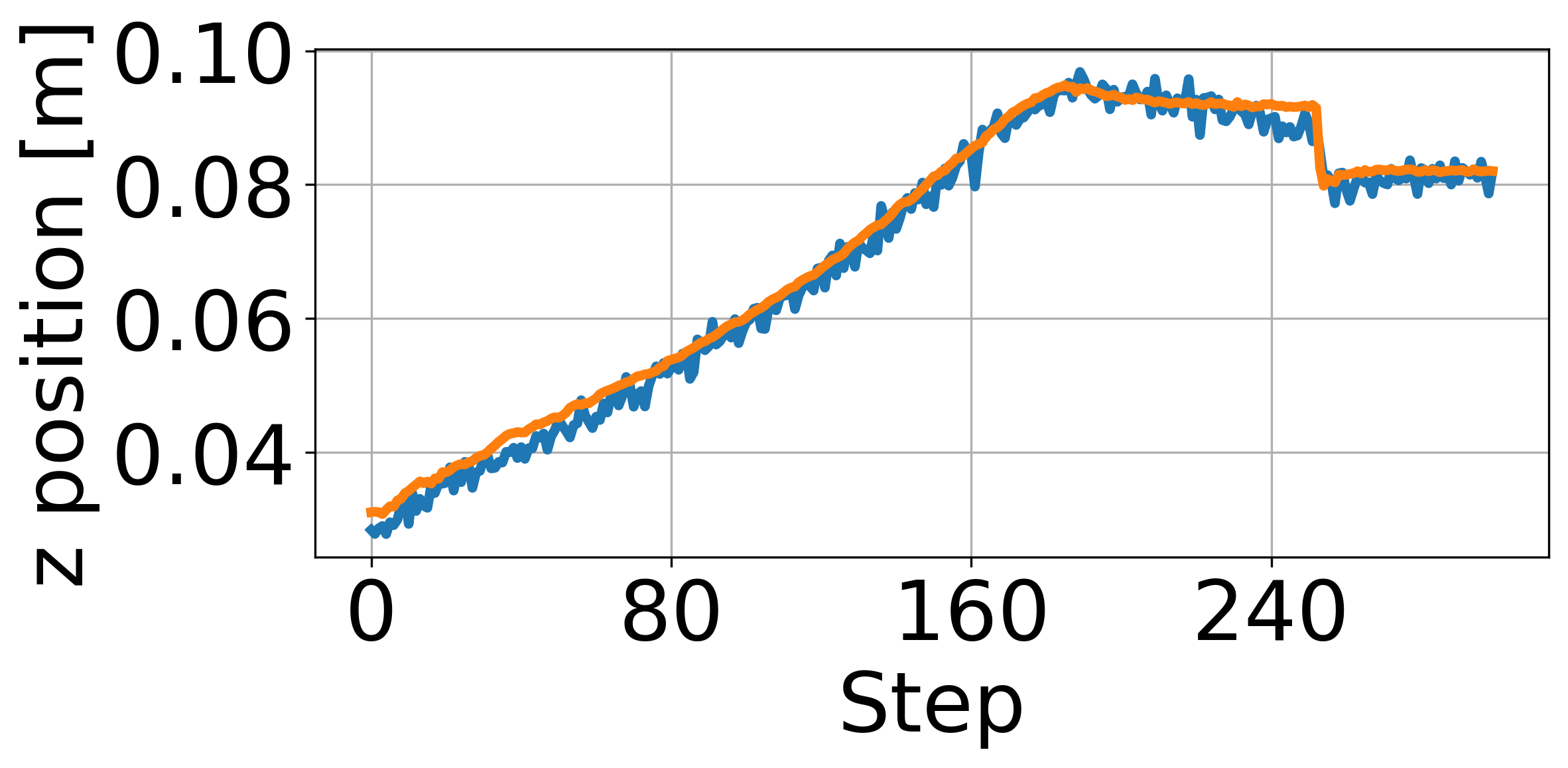}
    \end{subfigure}
            \begin{subfigure}{0.241\textwidth}
        \centering
        \includegraphics[width=1.0\linewidth]{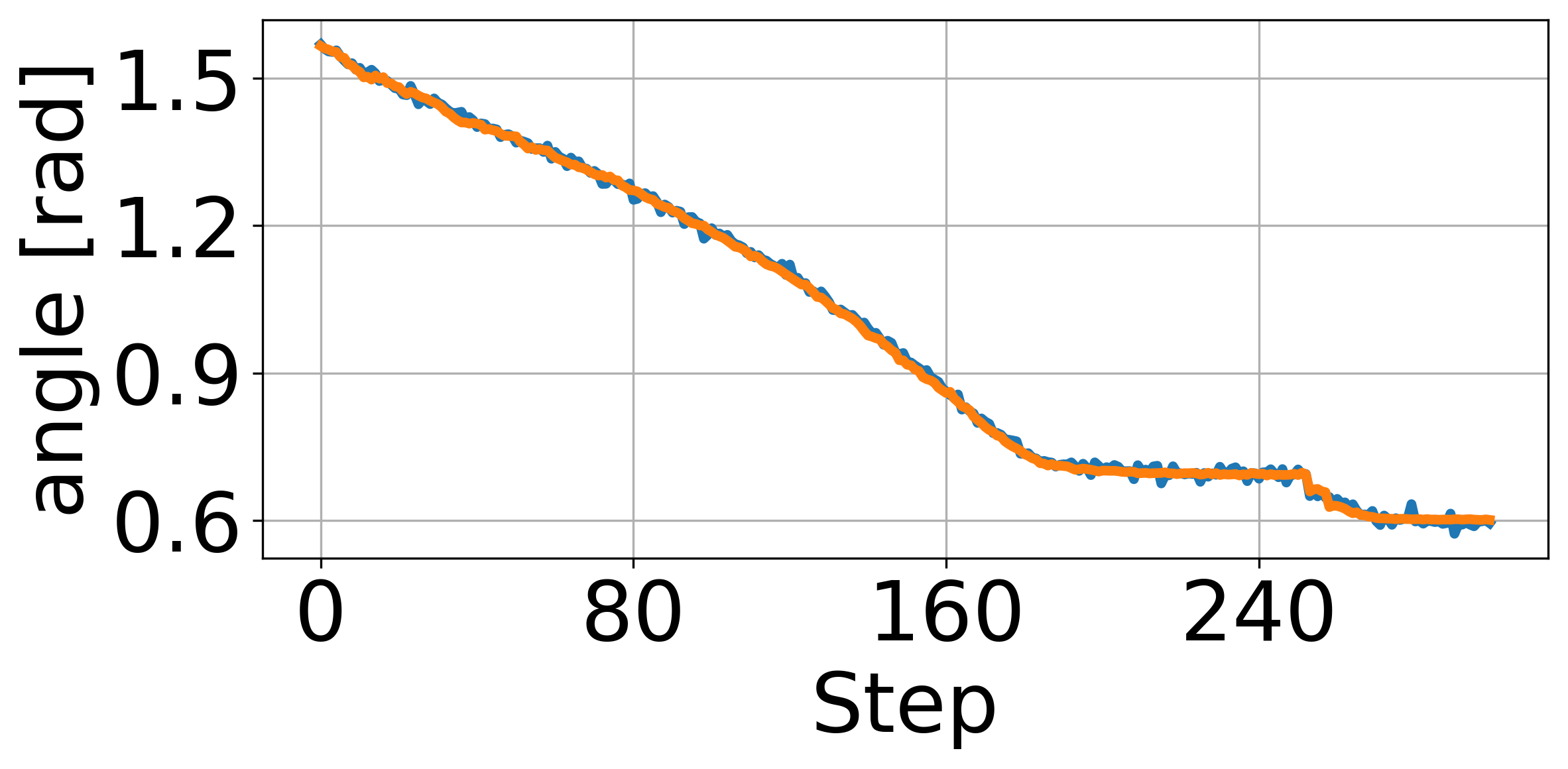}
    \end{subfigure}
        \hfill
    \caption{Comparison of our student estimator's predictions and the ground truth for the wall friction constant, $y$- and $z$-position of the object, and orientation along $x$-axis, for the pivoting with a wall.}
    \label{student-est-fig}
\end{figure}

\textbf{How accurately can the student estimator predict the privileged information?}
We deploy the trained student estimator and the teacher policy in MuJoCo and collect both the ground-truth privileged information and the corresponding student estimator's predictions. 
Representative results are shown in \fig{student-est-fig}, demonstrating that our student estimator can successfully predict the privileged information with reasonable accuracy.

\begin{figure}
    \centering
    \begin{subfigure}{0.495\textwidth}
        \centering
        \includegraphics[width=1.0\linewidth]{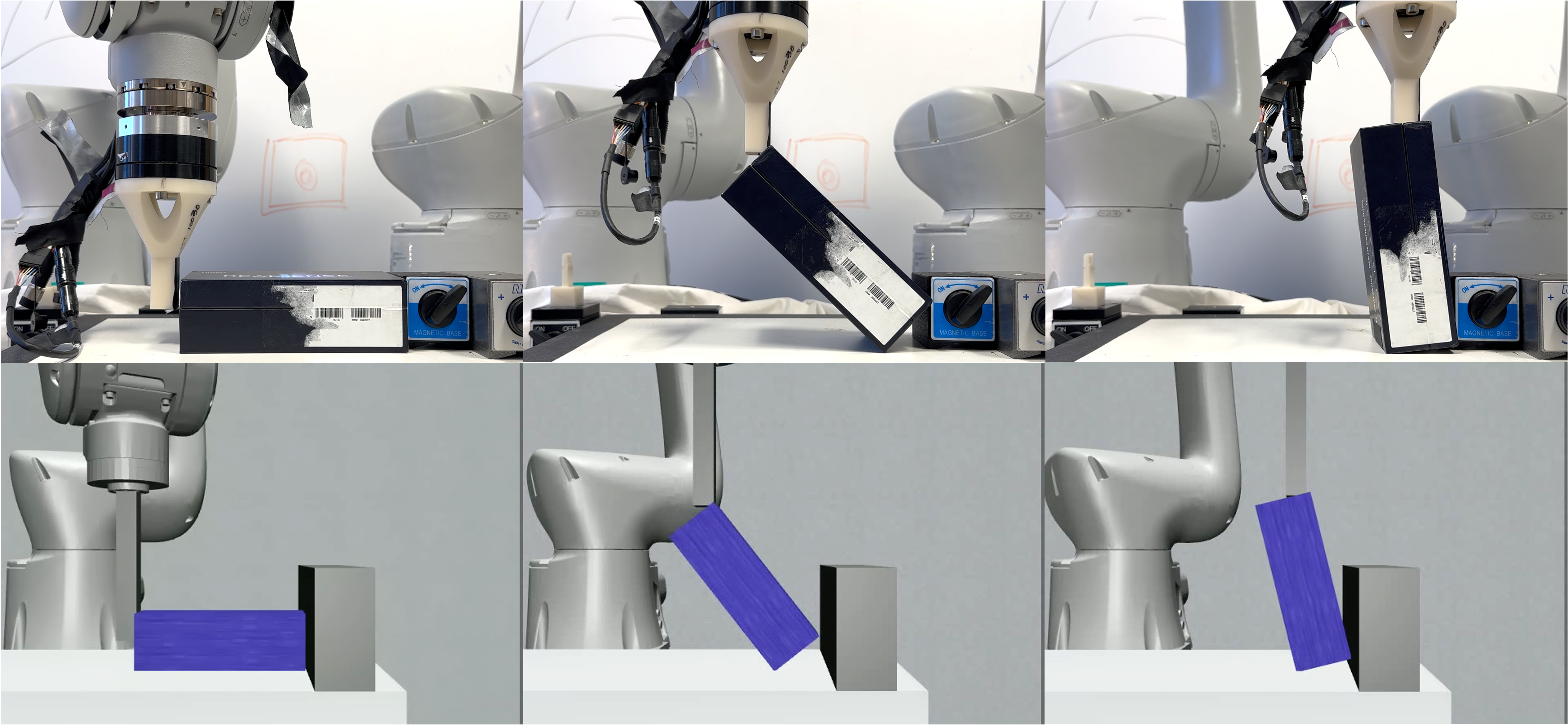}
        \caption{Pivoting with external wall}
    \end{subfigure}
    \begin{subfigure}{0.495\textwidth}
        \centering
        \includegraphics[width=1.0\linewidth]{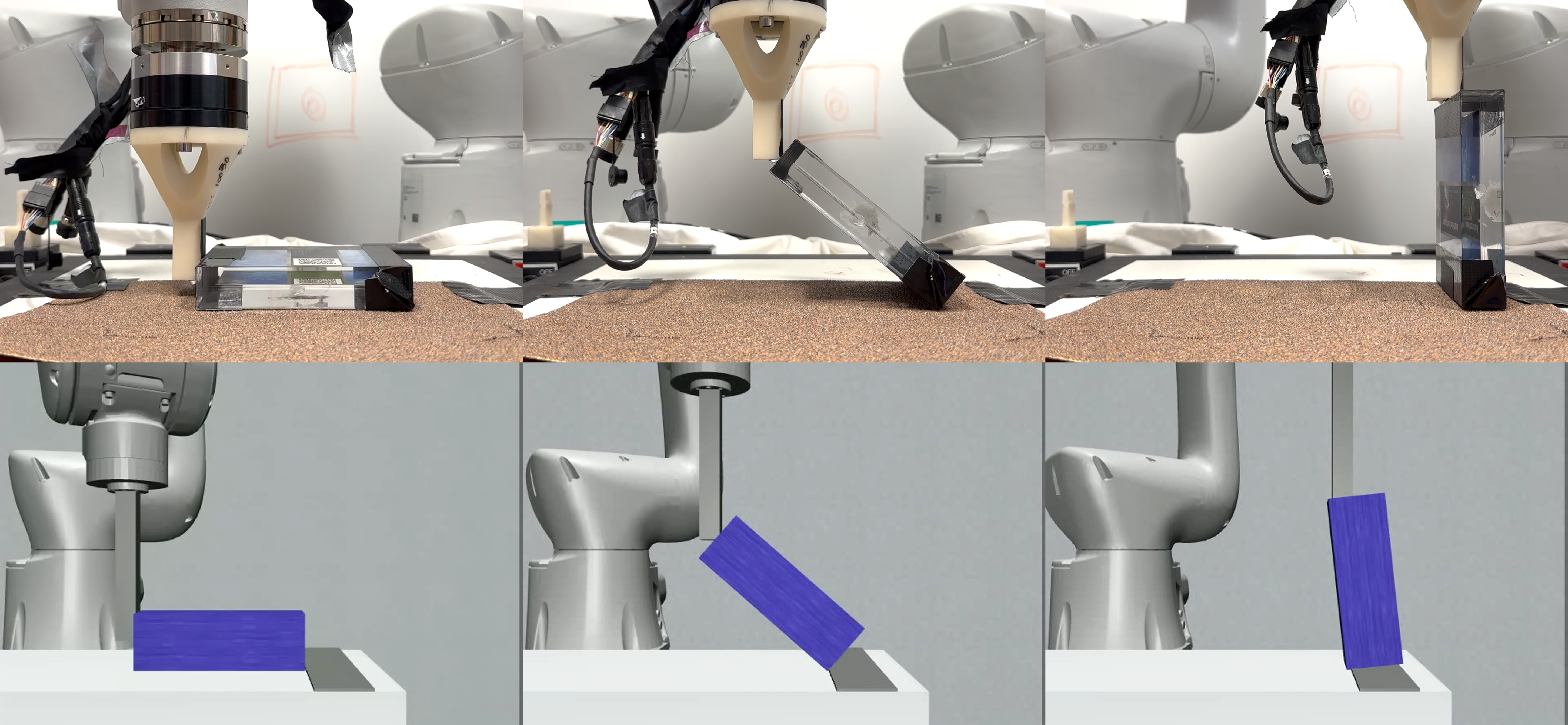}
\caption{Pivoting without external wall}
        \label{fig:hardware_fig_taskB}
    \end{subfigure}
        \hfill
    \caption{
    Snapshots of successful pivoting manipulation in simulation and real-world. All hardware experiment videos are found at \url{https://youtu.be/akjGDgfwLbM?si=0umnlPv-PMCTk_6F}. 
    }
    \label{hardware_fig}
\end{figure}

\textbf{Hardware Experiments}.
We deploy our teacher policy and student estimator on the real robot using zero-shot sim-to-real transfer. 
Overall, the policy successfully completes the desired task without access to privileged information as shown in \fig{hardware_fig}. 




\section{Conclusion}
\label{sec:conclusion}
	In this paper, we 
    present a framework for learning closed-loop controllers and estimators for contact-rich pivoting manipulation. 
    We first leverage CITO to generate high-quality demonstrations, including object and robot states, contact forces, and extrinsic contact location. Then, we perform demonstration-guided RL using these demonstrations for training a teacher policy, enabling sample-efficient learning. 
    Furthermore, we train a student estimator using only proprioception, vision, and force sensing, in order to predict the privileged information the teacher policy uses. 
    %
   Our framework is evaluated over several tasks, including the comparison against several baselines, and achieves successful zero-shot sim-to-real transfer in real-world experiments.




\section{Limitations}
\label{sec:limitations}

Our work has the following limitations.
First, we evaluate our framework exclusively on the pivoting task and do not demonstrate results for other non-prehensile manipulation tasks such as pushing and sliding. 
This choice was intentional to isolate and analyze key system components.
However, our method does not assume task-specific priors and is applicable to a broader range of non-prehensile tasks, as long as CITO can generate dynamically feasible demonstrations, which is possible via the approach in \cite{shirai2025hierarchical} or other CITO methods such as \cite{pang2023global}.
Extending the framework to a multi-task setup or evaluating generalizing across different manipulation tasks remains a promising direction for future work.


Second, all evaluations in this work are performed on convex objects (e.g., boxes), and we do not report results for non-convex geometries.
While none of our framework's modules rely on convexity assumptions, handling non-convex objects introduces additional complexity in contact reasoning. A promising future direction is to train both the teacher policy and student estimator over a distribution of object shapes, enabling generalization across object geometries.


Third, we assume that objects are rigid and non-articulated in this work. This limitation arises from the nature of CITO, which our method relies on for generating demonstrations. 
The CITO we used in this paper \cite{shirai2025hierarchical} or other CITO methods \cite{manchester2019variational, aceituno2020global} do not support such dynamics.
As a result, the performance may decrease when handling deformable or articulated objects. 
Extending CITO to support such dynamics, potentially through differentiable simulation, would be a valuable extension.


Fourth, we restrict our focus to quasistatic manipulation in SE(2), which limits the applicability of our proposed framework. 
Other CITO methods (e.g., \cite{aceituno2020global, pang2023global}) also rely on quasistatic manipulation model assumption.
Throughout this work, we rely on this assumption. In particular:
\begin{itemize}
    \item The model-based planner we used in this paper \cite{shirai2025hierarchical} operates in SE(2) manipulation due to its inability to model 3D contact dynamics. 
    \item We leverage the SE(2) assumption to facilitate sim2real transfer, using segmentation to simplify object pose estimation. 
\end{itemize}
To overcome the first limitation, the planner could be extended to support 3D contact dynamics. 
For the second, incorporating additional camera views to obtain segmentation masks from multiple angles would help lift the restriction to planar manipulation. 

Fifth, we empirically observe that policy learning becomes significantly more challenging when the range of domain randomization over table and wall coefficients increases. 
This is expected, as higher friction can lead to sticking contact while lower friction leads to sliding contact, resulting in multi-modal interaction behavior.
In such cases, a richer policy representation, beyond a basic MLP, may be necessary to achieve efficient learning. 


Finally, during real-world deployment, we occasionally observed slight object slip (i.e., incipient slip \cite{dong2019maintaining, shirai2023tactile, hogan2020tactile}) relative to the robot,  resulting in task failure. This issue is quite challenging: the 
 slip must be large enough to produce detectable changes in sensor signals (e.g., vision or force), allowing the student estimator to recognize it, yet small enough to avoid complete contact loss. 
This limitation is not a significant issue for other works focused on table-top manipulation \cite{pang2023global}, since objects are inherently stable. 
Addressing this limitation would likely require higher-resolution sensing or slip-specific estimation modules---for example, integrating visuotactile sensing (e.g., GelSight \cite{yuan2017gelsight}) or augmenting the student model with incipient slip prediction capabilities.




\clearpage


\bibliography{main}  

\begin{thebibliography}{77}
\providecommand{\natexlab}[1]{#1}
\providecommand{\url}[1]{\texttt{#1}}
\expandafter\ifx\csname urlstyle\endcsname\relax
  \providecommand{\doi}[1]{doi: #1}\else
  \providecommand{\doi}{doi: \begingroup \urlstyle{rm}\Url}\fi

\bibitem[Billard and Kragic(2019)]{billard2019trends}
A.~Billard and D.~Kragic.
\newblock Trends and challenges in robot manipulation.
\newblock \emph{Science}, 364\penalty0 (6446):\penalty0 eaat8414, 2019.

\bibitem[Mason(2018)]{mason2018toward}
M.~T. Mason.
\newblock Toward robotic manipulation.
\newblock \emph{Annual Review of Control, Robotics, and Autonomous Systems}, 1:\penalty0 1--28, 2018.

\bibitem[Rodriguez(2021)]{rodriguez2021unstable}
A.~Rodriguez.
\newblock The unstable queen: Uncertainty, mechanics, and tactile feedback.
\newblock \emph{Science Robotics}, 6\penalty0 (54):\penalty0 eabi4667, 2021.

\bibitem[Sleiman et~al.(2019)Sleiman, Carius, Grandia, Wermelinger, and Hutter]{sleiman2019contact}
J.-P. Sleiman, J.~Carius, R.~Grandia, M.~Wermelinger, and M.~Hutter.
\newblock Contact-implicit trajectory optimization for dynamic object manipulation.
\newblock In \emph{2019 IEEE/RSJ international conference on intelligent robots and systems (IROS)}, pages 6814--6821. IEEE, 2019.

\bibitem[Pang et~al.(2023)Pang, Suh, Yang, and Tedrake]{pang2023global}
T.~Pang, H.~T. Suh, L.~Yang, and R.~Tedrake.
\newblock Global planning for contact-rich manipulation via local smoothing of quasi-dynamic contact models.
\newblock \emph{IEEE Transactions on robotics}, 2023.

\bibitem[Le~Cleac'h et~al.(2024)Le~Cleac'h, Howell, Yang, Lee, Zhang, Bishop, Schwager, and Manchester]{le2024fast}
S.~Le~Cleac'h, T.~A. Howell, S.~Yang, C.-Y. Lee, J.~Zhang, A.~Bishop, M.~Schwager, and Z.~Manchester.
\newblock Fast contact-implicit model predictive control.
\newblock \emph{IEEE Transactions on Robotics}, 40:\penalty0 1617--1629, 2024.

\bibitem[Moura et~al.(2022)Moura, Stouraitis, and Vijayakumar]{moura2022non}
J.~Moura, T.~Stouraitis, and S.~Vijayakumar.
\newblock Non-prehensile planar manipulation via trajectory optimization with complementarity constraints.
\newblock In \emph{2022 International Conference on Robotics and Automation (ICRA)}, pages 970--976. IEEE, 2022.

\bibitem[Wijayarathne et~al.(2023)Wijayarathne, Zhou, Zhao, and Hammond]{wijayarathne2023real}
L.~Wijayarathne, Z.~Zhou, Y.~Zhao, and F.~L. Hammond.
\newblock Real-time deformable-contact-aware model predictive control for force-modulated manipulation.
\newblock \emph{IEEE Transactions on Robotics}, 39\penalty0 (5):\penalty0 3549--3566, 2023.

\bibitem[Shirai et~al.(2020)Shirai, Lin, Tanaka, Mehta, and Hong]{9113247}
Y.~Shirai, X.~Lin, Y.~Tanaka, A.~Mehta, and D.~Hong.
\newblock Risk-aware motion planning for a limbed robot with stochastic gripping forces using nonlinear programming.
\newblock \emph{IEEE Robotics and Automation Letters}, 5\penalty0 (4):\penalty0 4994--5001, 2020.
\newblock \doi{10.1109/LRA.2020.3001503}.

\bibitem[Levine et~al.(2016)Levine, Finn, Darrell, and Abbeel]{levine2016end}
S.~Levine, C.~Finn, T.~Darrell, and P.~Abbeel.
\newblock End-to-end training of deep visuomotor policies.
\newblock \emph{Journal of Machine Learning Research}, 17\penalty0 (39):\penalty0 1--40, 2016.

\bibitem[Rajeswaran et~al.(2018)Rajeswaran, Kumar, Gupta, Vezzani, Schulman, Todorov, and Levine]{Rajeswaran-RSS-18}
A.~Rajeswaran, V.~Kumar, A.~Gupta, G.~Vezzani, J.~Schulman, E.~Todorov, and S.~Levine.
\newblock Learning complex dexterous manipulation with deep reinforcement learning and demonstrations.
\newblock In \emph{Proceedings of Robotics: Science and Systems}, Pittsburgh, Pennsylvania, June 2018.
\newblock \doi{10.15607/RSS.2018.XIV.049}.

\bibitem[Beltran-Hernandez et~al.(2020)Beltran-Hernandez, Petit, Ramirez-Alpizar, and Harada]{beltran2020variable}
C.~C. Beltran-Hernandez, D.~Petit, I.~G. Ramirez-Alpizar, and K.~Harada.
\newblock Variable compliance control for robotic peg-in-hole assembly: A deep-reinforcement-learning approach.
\newblock \emph{Applied Sciences}, 10\penalty0 (19):\penalty0 6923, 2020.

\bibitem[Lee et~al.(2020)Lee, Hwangbo, Wellhausen, Koltun, and Hutter]{lee2020learning}
J.~Lee, J.~Hwangbo, L.~Wellhausen, V.~Koltun, and M.~Hutter.
\newblock Learning quadrupedal locomotion over challenging terrain.
\newblock \emph{Science robotics}, 5\penalty0 (47):\penalty0 eabc5986, 2020.

\bibitem[Andrychowicz et~al.(2020)Andrychowicz, Baker, Chociej, Jozefowicz, McGrew, Pachocki, Petron, Plappert, Powell, Ray, et~al.]{andrychowicz2020learning}
O.~M. Andrychowicz, B.~Baker, M.~Chociej, R.~Jozefowicz, B.~McGrew, J.~Pachocki, A.~Petron, M.~Plappert, G.~Powell, A.~Ray, et~al.
\newblock Learning dexterous in-hand manipulation.
\newblock \emph{The International Journal of Robotics Research}, 39\penalty0 (1):\penalty0 3--20, 2020.

\bibitem[Handa et~al.(2023)Handa, Allshire, Makoviychuk, Petrenko, Singh, Liu, Makoviichuk, Van~Wyk, Zhurkevich, Sundaralingam, et~al.]{handa2023dextreme}
A.~Handa, A.~Allshire, V.~Makoviychuk, A.~Petrenko, R.~Singh, J.~Liu, D.~Makoviichuk, K.~Van~Wyk, A.~Zhurkevich, B.~Sundaralingam, et~al.
\newblock Dextreme: Transfer of agile in-hand manipulation from simulation to reality.
\newblock In \emph{2023 IEEE International Conference on Robotics and Automation (ICRA)}, pages 5977--5984. IEEE, 2023.

\bibitem[Qi et~al.(2023)Qi, Yi, Suresh, Lambeta, Ma, Calandra, and Malik]{qi2023general}
H.~Qi, B.~Yi, S.~Suresh, M.~Lambeta, Y.~Ma, R.~Calandra, and J.~Malik.
\newblock General in-hand object rotation with vision and touch.
\newblock In \emph{Conference on Robot Learning}, pages 2549--2564. PMLR, 2023.

\bibitem[Chen et~al.(2020)Chen, Zhou, Koltun, and Kr{\"a}henb{\"u}hl]{chen2020learning}
D.~Chen, B.~Zhou, V.~Koltun, and P.~Kr{\"a}henb{\"u}hl.
\newblock Learning by cheating.
\newblock In \emph{Conference on robot learning}, pages 66--75. PMLR, 2020.

\bibitem[Aydinoglu et~al.(2024)Aydinoglu, Wei, Huang, and Posa]{aydinoglu2024consensus}
A.~Aydinoglu, A.~Wei, W.-C. Huang, and M.~Posa.
\newblock Consensus complementarity control for multi-contact mpc.
\newblock \emph{IEEE Transactions on Robotics}, 2024.

\bibitem[Aceituno-Cabezas and Rodriguez(2020)]{aceituno2020global}
B.~Aceituno-Cabezas and A.~Rodriguez.
\newblock A global quasi-dynamic model for contact-trajectory optimization in manipulation.
\newblock In \emph{Robotics: Science and Systems Foundation}, 2020.

\bibitem[Shirai et~al.(2024)Shirai, Jha, and Raghunathan]{shirai2024robust}
Y.~Shirai, D.~K. Jha, and A.~U. Raghunathan.
\newblock Robust pivoting manipulation using contact implicit bilevel optimization.
\newblock \emph{IEEE Transactions on Robotics}, 40:\penalty0 3425--3444, 2024.

\bibitem[Shirai et~al.(2022)Shirai, Jha, Raghunathan, and Romeres]{shirai2022robust}
Y.~Shirai, D.~K. Jha, A.~U. Raghunathan, and D.~Romeres.
\newblock Robust pivoting: Exploiting frictional stability using bilevel optimization.
\newblock In \emph{2022 International Conference on Robotics and Automation (ICRA)}, pages 992--998. IEEE, 2022.

\bibitem[Hogan et~al.(2020)Hogan, Ballester, Dong, and Rodriguez]{hogan2020tactile}
F.~R. Hogan, J.~Ballester, S.~Dong, and A.~Rodriguez.
\newblock Tactile dexterity: Manipulation primitives with tactile feedback.
\newblock In \emph{2020 IEEE international conference on robotics and automation (ICRA)}, pages 8863--8869. IEEE, 2020.

\bibitem[Jin et~al.(2021)Jin, Romeres, Ragunathan, Jha, and Tomizuka]{jin2021trajectory}
S.~Jin, D.~Romeres, A.~Ragunathan, D.~K. Jha, and M.~Tomizuka.
\newblock Trajectory optimization for manipulation of deformable objects: Assembly of belt drive units.
\newblock In \emph{2021 IEEE International Conference on Robotics and Automation (ICRA)}, pages 10002--10008. IEEE, 2021.

\bibitem[Shirai et~al.(2025)Shirai, Zhao, Suh, Zhu, Ni, Wang, Simchowitz, and Pang]{shirai2024linear}
Y.~Shirai, T.~Zhao, H.~Suh, H.~Zhu, X.~Ni, J.~Wang, M.~Simchowitz, and T.~Pang.
\newblock Is linear feedback on smoothed dynamics sufficient for stabilizing contact-rich plans?
\newblock \emph{2025 International Conference on Robotics and Automation (ICRA)}, 2025.

\bibitem[Hogan and Rodriguez(2020)]{hogan2020reactive}
F.~R. Hogan and A.~Rodriguez.
\newblock Reactive planar non-prehensile manipulation with hybrid model predictive control.
\newblock \emph{The International Journal of Robotics Research}, 39\penalty0 (7):\penalty0 755--773, 2020.

\bibitem[Shirai et~al.(2022)Shirai, Lin, Schperberg, Tanaka, Kato, Vichathorn, and Hong]{shirai_2022iros}
Y.~Shirai, X.~Lin, A.~Schperberg, Y.~Tanaka, H.~Kato, V.~Vichathorn, and D.~Hong.
\newblock Simultaneous contact-rich grasping and locomotion via distributed optimization enabling free-climbing for multi-limbed robots.
\newblock In \emph{Proc. 2022 IEEE/RSJ Int. Conf. Intell. Rob. Syst.}, pages 13563--13570, 2022.
\newblock \doi{10.1109/IROS47612.2022.9981579}.

\bibitem[Aydinoglu et~al.(2021)Aydinoglu, Sieg, Preciado, and Posa]{aydinoglu2021stabilization}
A.~Aydinoglu, P.~Sieg, V.~M. Preciado, and M.~Posa.
\newblock Stabilization of complementarity systems via contact-aware controllers.
\newblock \emph{IEEE Transactions on Robotics}, 38\penalty0 (3):\penalty0 1735--1754, 2021.

\bibitem[Gu et~al.(2017)Gu, Holly, Lillicrap, and Levine]{gu2017deep}
S.~Gu, E.~Holly, T.~Lillicrap, and S.~Levine.
\newblock Deep reinforcement learning for robotic manipulation with asynchronous off-policy updates.
\newblock In \emph{2017 IEEE international conference on robotics and automation (ICRA)}, pages 3389--3396. IEEE, 2017.

\bibitem[Chen et~al.(2023)Chen, Tippur, Wu, Kumar, Adelson, and Agrawal]{chen2023visual}
T.~Chen, M.~Tippur, S.~Wu, V.~Kumar, E.~Adelson, and P.~Agrawal.
\newblock Visual dexterity: In-hand reorientation of novel and complex object shapes.
\newblock \emph{Science Robotics}, 8\penalty0 (84):\penalty0 eadc9244, 2023.

\bibitem[Chi et~al.(2023)Chi, Xu, Feng, Cousineau, Du, Burchfiel, Tedrake, and Song]{chi2023diffusion}
C.~Chi, Z.~Xu, S.~Feng, E.~Cousineau, Y.~Du, B.~Burchfiel, R.~Tedrake, and S.~Song.
\newblock Diffusion policy: Visuomotor policy learning via action diffusion.
\newblock \emph{The International Journal of Robotics Research}, page 02783649241273668, 2023.

\bibitem[Fu et~al.(2024)Fu, Zhao, and Finn]{fu2024mobile}
Z.~Fu, T.~Z. Zhao, and C.~Finn.
\newblock Mobile aloha: Learning bimanual mobile manipulation with low-cost whole-body teleoperation.
\newblock \emph{arXiv preprint arXiv:2401.02117}, 2024.

\bibitem[Black et~al.(2024)Black, Brown, Driess, Esmail, Equi, Finn, Fusai, Groom, Hausman, Ichter, et~al.]{black2024pi_0}
K.~Black, N.~Brown, D.~Driess, A.~Esmail, M.~Equi, C.~Finn, N.~Fusai, L.~Groom, K.~Hausman, B.~Ichter, et~al.
\newblock $\pi_0$: A vision-language-action flow model for general robot control.
\newblock \emph{arXiv preprint arXiv:2410.24164}, 2024.

\bibitem[Team et~al.(2025)Team, Abeyruwan, Ainslie, Alayrac, Arenas, Armstrong, Balakrishna, Baruch, Bauza, Blokzijl, et~al.]{team2025gemini}
G.~R. Team, S.~Abeyruwan, J.~Ainslie, J.-B. Alayrac, M.~G. Arenas, T.~Armstrong, A.~Balakrishna, R.~Baruch, M.~Bauza, M.~Blokzijl, et~al.
\newblock Gemini robotics: Bringing ai into the physical world.
\newblock \emph{arXiv preprint arXiv:2503.20020}, 2025.

\bibitem[Xu et~al.(2025)Xu, Uppuluri, Zhang, Fitch, Crandall, Shou, Wang, and She]{xu2025unit}
Z.~Xu, R.~Uppuluri, X.~Zhang, C.~Fitch, P.~G. Crandall, W.~Shou, D.~Wang, and Y.~She.
\newblock Unit: Data efficient tactile representation with generalization to unseen objects.
\newblock \emph{IEEE Robotics and Automation Letters}, 2025.

\bibitem[Lin et~al.(2025)Lin, Sachdev, Fan, Malik, and Zhu]{lin2025sim}
T.~Lin, K.~Sachdev, L.~Fan, J.~Malik, and Y.~Zhu.
\newblock Sim-to-real reinforcement learning for vision-based dexterous manipulation on humanoids.
\newblock \emph{arXiv preprint arXiv:2502.20396}, 2025.

\bibitem[Noseworthy et~al.(2025)Noseworthy, Tang, Wen, Handa, Kessens, Roy, Fox, Ramos, Narang, and Akinola]{noseworthy2025forge}
M.~Noseworthy, B.~Tang, B.~Wen, A.~Handa, C.~Kessens, N.~Roy, D.~Fox, F.~Ramos, Y.~Narang, and I.~Akinola.
\newblock Forge: Force-guided exploration for robust contact-rich manipulation under uncertainty.
\newblock \emph{IEEE Robotics and Automation Letters}, 2025.

\bibitem[Seo et~al.(2023)Seo, Han, Sim, Bang, Gonzalez, Sentis, and Zhu]{seo2023deep}
M.~Seo, S.~Han, K.~Sim, S.~H. Bang, C.~Gonzalez, L.~Sentis, and Y.~Zhu.
\newblock Deep imitation learning for humanoid loco-manipulation through human teleoperation.
\newblock In \emph{2023 IEEE-RAS 22nd International Conference on Humanoid Robots (Humanoids)}, pages 1--8. IEEE, 2023.

\bibitem[Vecerik et~al.(2017)Vecerik, Hester, Scholz, Wang, Pietquin, Piot, Heess, Roth{\"o}rl, Lampe, and Riedmiller]{vecerik2017leveraging}
M.~Vecerik, T.~Hester, J.~Scholz, F.~Wang, O.~Pietquin, B.~Piot, N.~Heess, T.~Roth{\"o}rl, T.~Lampe, and M.~Riedmiller.
\newblock Leveraging demonstrations for deep reinforcement learning on robotics problems with sparse rewards.
\newblock \emph{arXiv preprint arXiv:1707.08817}, 2017.

\bibitem[Ankile et~al.(2024)Ankile, Simeonov, Shenfeld, Torne, and Agrawal]{ankile2024imitation}
L.~Ankile, A.~Simeonov, I.~Shenfeld, M.~Torne, and P.~Agrawal.
\newblock From imitation to refinement--residual rl for precise assembly.
\newblock \emph{arXiv preprint arXiv:2407.16677}, 2024.

\bibitem[Ota et~al.(2021)Ota, Jha, Onishi, Kanezaki, Yoshiyasu, Sasaki, Mariyama, and Nikovski]{ota2021deep}
K.~Ota, D.~Jha, T.~Onishi, A.~Kanezaki, Y.~Yoshiyasu, Y.~Sasaki, T.~Mariyama, and D.~Nikovski.
\newblock Deep reactive planning in dynamic environments.
\newblock In \emph{Conference on Robot Learning}, pages 1943--1957. PMLR, 2021.

\bibitem[Xiong et~al.(2021)Xiong, Li, Chen, Bharadhwaj, Sinha, and Garg]{xiong2021learning}
H.~Xiong, Q.~Li, Y.-C. Chen, H.~Bharadhwaj, S.~Sinha, and A.~Garg.
\newblock Learning by watching: Physical imitation of manipulation skills from human videos.
\newblock In \emph{2021 IEEE/RSJ International Conference on Intelligent Robots and Systems (IROS)}, pages 7827--7834. IEEE, 2021.

\bibitem[Zhang et~al.(2018)Zhang, McCarthy, Jow, Lee, Chen, Goldberg, and Abbeel]{zhang2018deep}
T.~Zhang, Z.~McCarthy, O.~Jow, D.~Lee, X.~Chen, K.~Goldberg, and P.~Abbeel.
\newblock Deep imitation learning for complex manipulation tasks from virtual reality teleoperation.
\newblock In \emph{2018 IEEE international conference on robotics and automation (ICRA)}, pages 5628--5635. Ieee, 2018.

\bibitem[Fuchioka et~al.(2023)Fuchioka, Xie, and Van~de Panne]{fuchioka2023opt}
Y.~Fuchioka, Z.~Xie, and M.~Van~de Panne.
\newblock Opt-mimic: Imitation of optimized trajectories for dynamic quadruped behaviors.
\newblock In \emph{2023 IEEE International Conference on Robotics and Automation (ICRA)}, pages 5092--5098. IEEE, 2023.

\bibitem[Sleiman et~al.(2024)Sleiman, Mittal, and Hutter]{sleiman2024guided}
J.-P. Sleiman, M.~Mittal, and M.~Hutter.
\newblock Guided reinforcement learning for robust multi-contact loco-manipulation.
\newblock In \emph{8th Annual Conference on Robot Learning (CoRL 2024)}, 2024.

\bibitem[Bruedigam et~al.(2025)Bruedigam, Abbas, Sorokin, Fang, Hung, Guru, Sosnowski, Wang, Hirche, and Cleac'h]{pmlr-v270-bruedigam25a}
J.~Bruedigam, A.~A. Abbas, M.~Sorokin, K.~Fang, B.~Hung, M.~Guru, S.~G. Sosnowski, J.~Wang, S.~Hirche, and S.~L. Cleac'h.
\newblock Jacta: A versatile planner for learning dexterous and whole-body manipulation.
\newblock In P.~Agrawal, O.~Kroemer, and W.~Burgard, editors, \emph{Proceedings of The 8th Conference on Robot Learning}, volume 270 of \emph{Proceedings of Machine Learning Research}, pages 994--1020. PMLR, 06--09 Nov 2025.
\newblock URL \url{https://proceedings.mlr.press/v270/bruedigam25a.html}.

\bibitem[Hou et~al.(2024)Hou, Liu, Chi, Cousineau, Kuppuswamy, Feng, Burchfiel, and Song]{hou2024adaptive}
Y.~Hou, Z.~Liu, C.~Chi, E.~Cousineau, N.~Kuppuswamy, S.~Feng, B.~Burchfiel, and S.~Song.
\newblock Adaptive compliance policy: Learning approximate compliance for diffusion guided control.
\newblock \emph{arXiv preprint arXiv:2410.09309}, 2024.

\bibitem[Chen et~al.(2025)Chen, Yu, Choi, Cutkosky, and Bohg]{chen2025dexforce}
C.~Chen, Z.~Yu, H.~Choi, M.~Cutkosky, and J.~Bohg.
\newblock Dexforce: Extracting force-informed actions from kinesthetic demonstrations for dexterous manipulation.
\newblock \emph{arXiv preprint arXiv:2501.10356}, 2025.

\bibitem[Olson(2011)]{5979561}
E.~Olson.
\newblock Apriltag: A robust and flexible visual fiducial system.
\newblock In \emph{2011 IEEE International Conference on Robotics and Automation}, pages 3400--3407, 2011.
\newblock \doi{10.1109/ICRA.2011.5979561}.

\bibitem[Ferrandis et~al.(2024)Ferrandis, Moura, and Vijayakumar]{ferrandis2024learning}
J.~D.~A. Ferrandis, J.~Moura, and S.~Vijayakumar.
\newblock Learning visuotactile estimation and control for non-prehensile manipulation under occlusions.
\newblock In \emph{8th Annual Conference on Robot Learning}, 2024.
\newblock URL \url{https://openreview.net/forum?id=oSU7M7MK6B}.

\bibitem[Kumar et~al.(2021)Kumar, Fu, Pathak, and Malik]{kumar2021rma}
A.~Kumar, Z.~Fu, D.~Pathak, and J.~Malik.
\newblock Rma: Rapid motor adaptation for legged robots.
\newblock 2021.

\bibitem[Miki et~al.(2022)Miki, Lee, Hwangbo, Wellhausen, Koltun, and Hutter]{miki2022learning}
T.~Miki, J.~Lee, J.~Hwangbo, L.~Wellhausen, V.~Koltun, and M.~Hutter.
\newblock Learning robust perceptive locomotion for quadrupedal robots in the wild.
\newblock \emph{Science robotics}, 7\penalty0 (62):\penalty0 eabk2822, 2022.

\bibitem[Kaufmann et~al.(2023)Kaufmann, Bauersfeld, Loquercio, M{\"u}ller, Koltun, and Scaramuzza]{kaufmann2023champion}
E.~Kaufmann, L.~Bauersfeld, A.~Loquercio, M.~M{\"u}ller, V.~Koltun, and D.~Scaramuzza.
\newblock Champion-level drone racing using deep reinforcement learning.
\newblock \emph{Nature}, 620\penalty0 (7976):\penalty0 982--987, 2023.

\bibitem[Su et~al.(2024)Su, Jia, Qin, Zhou, Macaluso, Huang, and Wang]{su2024sim2real}
E.~Su, C.~Jia, Y.~Qin, W.~Zhou, A.~Macaluso, B.~Huang, and X.~Wang.
\newblock Sim2real manipulation on unknown objects with tactile-based reinforcement learning.
\newblock In \emph{2024 IEEE International Conference on Robotics and Automation (ICRA)}, pages 9234--9241. IEEE, 2024.

\bibitem[Bauza et~al.(2024)Bauza, Chen, Dalibard, Gileadi, Hafner, Martins, Moore, Pevceviciute, Laurens, Rao, et~al.]{bauza2024demostart}
M.~Bauza, J.~E. Chen, V.~Dalibard, N.~Gileadi, R.~Hafner, M.~F. Martins, J.~Moore, R.~Pevceviciute, A.~Laurens, D.~Rao, et~al.
\newblock Demostart: Demonstration-led auto-curriculum applied to sim-to-real with multi-fingered robots.
\newblock \emph{arXiv preprint arXiv:2409.06613}, 2024.

\bibitem[Jiang et~al.(2024)Jiang, Wang, Zhang, Wu, and Fei-Fei]{jiang2024transic}
Y.~Jiang, C.~Wang, R.~Zhang, J.~Wu, and L.~Fei-Fei.
\newblock Transic: Sim-to-real policy transfer by learning from online correction.
\newblock In \emph{Conference on Robot Learning}, 2024.

\bibitem[Fuchioka et~al.(2024)Fuchioka, Beltran-Hernandez, Nguyen, and Hamaya]{fuchioka2024robotic}
Y.~Fuchioka, C.~C. Beltran-Hernandez, H.~Nguyen, and M.~Hamaya.
\newblock Robotic object insertion with a soft wrist through sim-to-real privileged training.
\newblock In \emph{2024 IEEE/RSJ International Conference on Intelligent Robots and Systems (IROS)}, pages 9159--9166. IEEE, 2024.

\bibitem[Shirai et~al.(2025)Shirai, Raghunathan, and Jha]{shirai2025hierarchical}
Y.~Shirai, A.~Raghunathan, and D.~K. Jha.
\newblock Hierarchical contact-rich trajectory optimization for multi-modal manipulation using tight convex relaxations.
\newblock \emph{2025 IEEE International Conference on Robotics and Automation}, 2025.

\bibitem[{Gurobi Optimization, LLC}(2024)]{gurobi}
{Gurobi Optimization, LLC}.
\newblock {Gurobi Optimizer Reference Manual}, 2024.
\newblock URL \url{https://www.gurobi.com}.

\bibitem[Gill et~al.(2005)Gill, Murray, and Saunders]{gill2005snopt}
P.~E. Gill, W.~Murray, and M.~A. Saunders.
\newblock Snopt: An sqp algorithm for large-scale constrained optimization.
\newblock \emph{SIAM review}, 47\penalty0 (1):\penalty0 99--131, 2005.

\bibitem[Khatib(1987)]{1087068}
O.~Khatib.
\newblock A unified approach for motion and force control of robot manipulators: The operational space formulation.
\newblock \emph{IEEE Journal on Robotics and Automation}, 3\penalty0 (1):\penalty0 43--53, 1987.
\newblock \doi{10.1109/JRA.1987.1087068}.

\bibitem[Zhang et~al.(2023{\natexlab{a}})Zhang, Jain, Huang, Tomizuka, and Romeres]{zhang2023learning}
X.~Zhang, S.~Jain, B.~Huang, M.~Tomizuka, and D.~Romeres.
\newblock Learning generalizable pivoting skills.
\newblock In \emph{2023 IEEE International Conference on Robotics and Automation (ICRA)}, pages 5865--5871. IEEE, 2023{\natexlab{a}}.

\bibitem[Zhang et~al.(2023{\natexlab{b}})Zhang, Wang, Sun, Wu, Zhu, and Tomizuka]{zhang2023efficient}
X.~Zhang, C.~Wang, L.~Sun, Z.~Wu, X.~Zhu, and M.~Tomizuka.
\newblock Efficient sim-to-real transfer of contact-rich manipulation skills with online admittance residual learning.
\newblock In \emph{Conference on Robot Learning}, pages 1621--1639. PMLR, 2023{\natexlab{b}}.

\bibitem[Tobin et~al.(2017)Tobin, Fong, Ray, Schneider, Zaremba, and Abbeel]{tobin2017domain}
J.~Tobin, R.~Fong, A.~Ray, J.~Schneider, W.~Zaremba, and P.~Abbeel.
\newblock Domain randomization for transferring deep neural networks from simulation to the real world.
\newblock In \emph{2017 IEEE/RSJ International Conference on Intelligent Robots and Systems (IROS)}, pages 23--30, 2017.
\newblock \doi{10.1109/IROS.2017.8202133}.

\bibitem[Bai et~al.(2018)Bai, Kolter, and Koltun]{bai2018empirical}
S.~Bai, J.~Z. Kolter, and V.~Koltun.
\newblock An empirical evaluation of generic convolutional and recurrent networks for sequence modeling.
\newblock \emph{arXiv preprint arXiv:1803.01271}, 2018.

\bibitem[Qi et~al.(2023)Qi, Kumar, Calandra, Ma, and Malik]{qi2023hand}
H.~Qi, A.~Kumar, R.~Calandra, Y.~Ma, and J.~Malik.
\newblock In-hand object rotation via rapid motor adaptation.
\newblock In \emph{Conference on Robot Learning}, pages 1722--1732. PMLR, 2023.

\bibitem[Zakka et~al.(2025)Zakka, Tabanpour, Liao, Haiderbhai, Holt, Luo, Allshire, Frey, Sreenath, Kahrs, et~al.]{zakka2025mujoco}
K.~Zakka, B.~Tabanpour, Q.~Liao, M.~Haiderbhai, S.~Holt, J.~Y. Luo, A.~Allshire, E.~Frey, K.~Sreenath, L.~A. Kahrs, et~al.
\newblock Mujoco playground.
\newblock \emph{arXiv preprint arXiv:2502.08844}, 2025.

\bibitem[Todorov et~al.(2012)Todorov, Erez, and Tassa]{todorov2012mujoco}
E.~Todorov, T.~Erez, and Y.~Tassa.
\newblock Mujoco: A physics engine for model-based control.
\newblock In \emph{2012 IEEE/RSJ International Conference on Intelligent Robots and Systems}, pages 5026--5033. IEEE, 2012.
\newblock \doi{10.1109/IROS.2012.6386109}.

\bibitem[Zhu et~al.(2020)Zhu, Wong, Mandlekar, Mart\'{i}n-Mart\'{i}n, Joshi, Nasiriany, Zhu, and Lin]{robosuite2020}
Y.~Zhu, J.~Wong, A.~Mandlekar, R.~Mart\'{i}n-Mart\'{i}n, A.~Joshi, S.~Nasiriany, Y.~Zhu, and K.~Lin.
\newblock robosuite: A modular simulation framework and benchmark for robot learning.
\newblock In \emph{arXiv preprint arXiv:2009.12293}, 2020.

\bibitem[Haarnoja et~al.(2018)Haarnoja, Zhou, Abbeel, and Levine]{haarnoja2018soft}
T.~Haarnoja, A.~Zhou, P.~Abbeel, and S.~Levine.
\newblock Soft actor-critic: Off-policy maximum entropy deep reinforcement learning with a stochastic actor.
\newblock In \emph{International conference on machine learning}, pages 1861--1870. Pmlr, 2018.

\bibitem[Ota(2020)]{ota2020tf2rl}
K.~Ota.
\newblock Tf2rl.
\newblock \url{https://github.com/keiohta/tf2rl/}, 2020.

\bibitem[mit()]{mitsubishielectricFactoryAutomation}
{F}actory {A}utomation - {M}itsubishi {E}lectric {A}mericas --- us.mitsubishielectric.com.
\newblock \url{https://us.mitsubishielectric.com/fa/en/products/rbt/collaborative-robot/}.
\newblock [Accessed 19-04-2025].

\bibitem[Zhao et~al.(2023)Zhao, Ding, An, Du, Yu, Li, Tang, and Wang]{zhao2023fast}
X.~Zhao, W.~Ding, Y.~An, Y.~Du, T.~Yu, M.~Li, M.~Tang, and J.~Wang.
\newblock Fast segment anything.
\newblock \emph{arXiv preprint arXiv:2306.12156}, 2023.

\bibitem[int()]{intelrealsenseDepthCamera}
{D}epth {C}amera {D}435 --- intelrealsense.com.
\newblock \url{https://www.intelrealsense.com/depth-camera-d435/}.
\newblock [Accessed 23-04-2025].

\bibitem[Manchester and Kuindersma(2019)]{manchester2019variational}
Z.~Manchester and S.~Kuindersma.
\newblock Variational contact-implicit trajectory optimization.
\newblock In \emph{Robotics Research: The 18th International Symposium ISRR}, pages 985--1000. Springer, 2019.

\bibitem[Dong et~al.(2019)Dong, Ma, Donlon, and Rodriguez]{dong2019maintaining}
S.~Dong, D.~Ma, E.~Donlon, and A.~Rodriguez.
\newblock Maintaining grasps within slipping bounds by monitoring incipient slip.
\newblock In \emph{2019 International Conference on Robotics and Automation (ICRA)}, pages 3818--3824. IEEE, 2019.

\bibitem[Shirai et~al.(2023)Shirai, Jha, Raghunathan, and Hong]{shirai2023tactile}
Y.~Shirai, D.~K. Jha, A.~U. Raghunathan, and D.~Hong.
\newblock Tactile tool manipulation.
\newblock In \emph{2023 IEEE International Conference on Robotics and Automation (ICRA)}, pages 12597--12603. IEEE, 2023.

\bibitem[Yuan et~al.(2017)Yuan, Dong, and Adelson]{yuan2017gelsight}
W.~Yuan, S.~Dong, and E.~H. Adelson.
\newblock Gelsight: High-resolution robot tactile sensors for estimating geometry and force.
\newblock \emph{Sensors}, 17\penalty0 (12):\penalty0 2762, 2017.

\end{thebibliography}

\clearpage
\appendix{}

  {\textbf{\LARGE Appendix}}

\section{CITO Details}

In this work, we use the CITO \eq{trajopt}, as presented in \cite{shirai2025hierarchical}. Given a task description, defined by the initial and goal poses in SE(2), along with privileged information (e.g., object mass, friction, and size, environment friction), the optimization problem in \eq{trajopt} is solved through a sequence of three optimization problems. 
The first optimization problem is as follows.
\begin{equation}\label{kine_eq}
\begin{aligned}
\min_{\bar{\mathbf{q}}_t^o,  \dot{\bar{\mathbf{q}}}_{t}^o,} &  \sum_{t=0}^{T} \|\bar{\mathbf{q}}^o_{t} - \bar{\mathbf{q}}_t^{o,\text{ref}}\|^2_{Q}\\
\text{s. t. }, \, &  h_1\left(\bar{\mathbf{q}}_t^o, \bar{\mathbf{q}}_{t+1}^o,  \dot{\bar{\mathbf{q}}}_{t}^o\right)  = \mathbf{0},  
\end{aligned}
\end{equation}
where $h_1$ is the set of constraints, including velocity constraints, bounds on variables, and signed distance function-based constraints to ensure collision avoidance between the object and the environment. 
The optimization problem in \eq{kine_eq} is used to obtain a kinematically feasible object pose trajectory and the corresponding extrinsic contact trajectory between the object and the environment. The optimization problem in \eq{kine_eq} is solved using SNOPT \cite{gill2005snopt}.

Second, after fixing the object pose trajectory $\bar{\mathbf{q}}_t^o$ to the solution obtained in the first stage, the following optimization problem is formulated to account for non-smooth constraints due to contact dynamics:
\begin{equation}\label{cont_eq}
\begin{aligned}
\text{Find } \ & \bar{\mathbf{q}}_t^r,  \dot{\bar{\mathbf{q}}}_{t}^r, \bar{\mathbf{y}}_t\\
\text{s. t. }, \, &  h_2\left(\bar{\mathbf{q}}_t^r, \bar{\mathbf{q}}_{t+1}^r,  \dot{\bar{\mathbf{q}}}_{t}^r, \bar{\mathbf{y}}_t\right)  = \mathbf{0}, 
\end{aligned}
\end{equation}
where $h_2$ represents the set of constraints used for considering non-smooth constraints, including contact making/breaking constraints, linearized force and moment balance constraints, and friction cone constraints. 
By solving \eq{cont_eq}, we obtain the object and robot trajectories that are not only kinematically feasible but also respect non-smooth contact constraints under linearized quasistatic dynamics. This optimization problem is a mixed-integer linear problem, which is efficiently solved using Gurobi \cite{gurobi}.

Finally, given the solution obtained from \eq{cont_eq}, we consider the following optimization problem.
\begin{equation}\label{dyn_trajopt_eq}
\begin{aligned}
\text{Find } \ & \bar{\mathbf{q}}_t,  \dot{\bar{\mathbf{q}}}_{t}, \bar{\mathbf{y}}_t\\
\text{s. t. }, \, &  h_3\left(\bar{\mathbf{q}}_t^r, \bar{\mathbf{q}}_{t+1}^r,  \dot{\bar{\mathbf{q}}}_{t}^r, \bar{\mathbf{y}}_t\right)  = \mathbf{0}, 
\end{aligned}
\end{equation}
where $h_3$ includes non-smooth sticking-sliding contact constraints using complementarity constraints as well as the original (not linearized) force and moment balance constraints. During solving \eq{dyn_trajopt_eq} the robot's positions are locally adjusted to satisfy the nonlinear force and moment balance constraints and sticking-sliding complementarity constraints. 
This optimization problem is solved through SNOPT. 
Note that, for certain combinations of dynamics parameters (e.g., mass, friction), the solver may return an infeasible solution. In such cases, we do not include these infeasible solutions in the demonstration dataset. 

It is worth noting that the solution obtained by sequentially solving the three optimization problems described above satisfies the full dynamics function \( f_\text{dyn} \) in \eq{trajopt} and is referred to as \emph{dynamically feasible}. In contrast, if we solve the same sequence of optimization problems while removing all constraints involving contact forces—such as force and moment balance constraints and friction cone constraints—the resulting solution is referred to as \emph{kinematically feasible} and satisfies the relaxed dynamics function \( f_\text{kin} \).

In summary, solving \eq{trajopt} involves a sequence of the three optimization problems described above, allowing for efficient computation by decoupling different sets of constraints across the subproblems. 
%
See \cite{shirai2025hierarchical} for more details. Finally, we summarize the parameters used in the above optimization problems in \tab{tab:cito-parameter}.

\begin{table}[h]
    \centering
    \caption{Hyperparameter setup for student estimator.}
    \begin{tabular}{cc} \toprule
           Parameter & Value \\ \midrule
         Optimizer & SNOPT for \eq{kine_eq} and \eq{dyn_trajopt_eq} and Gurobi for \eq{cont_eq}   \\
         $T$ & 60 for pivoting with-wall task and 150 for without-wall task\\
         time interval for integration & 0.1 s\\
         \bottomrule
    \end{tabular}
    \label{tab:cito-parameter}
\end{table}

\section{Training Details in Simulation}
In this section, we provide implementation details for training the teacher policy. 
The simulation environment is built using MuJoCo~\cite{todorov2012mujoco} with robosuite framework~\cite{robosuite2020}. We use Soft Actor Critic (SAC)~\cite{haarnoja2018soft} to train the teacher policy. The training parameters are summarized in \tab{tab:teacher-student-parameter}.

\begin{table}[h]
    \centering
    \caption{Hyperparameter setup for the teacher policy. Note that $\alpha_{i\in[1, \cdots, 8]}$ are the coefficients of the reward terms used for reward computation in \eq{reward_eq}.}
    \begin{tabular}{cc} \toprule
           Parameter & Value \\ \midrule
        total \# of steps &  300k for pivoting with-wall task and 1500k for without-wall task  \\
        batch size  & 4096 \\
        max \# of  step for timeout   & 300 \\
        Networks  & [128, 128] MLP \\
        learning rate for policy  & 1e-4 \\
        learning rate for Q function  & 3e-4 \\
        discount factor & 0.9 \\
        replay buffer size & 1e6 \\
        \# of episodes for evaluation  & 50 \\
        \# of episodes for warmstart & 50k \\
        $[\alpha_1, \alpha_2, \alpha_3, \alpha_4, \alpha_5, \alpha_6, \alpha_7, \alpha_8]$ & [1, 0.075, 10, -1, -50, -50, -0.005, 5] \\
         \bottomrule
    \end{tabular}
    \label{tab:teacher-student-parameter}
\end{table}

\begin{figure}[h]
    \centering
    \includegraphics[width=0.5\textwidth]{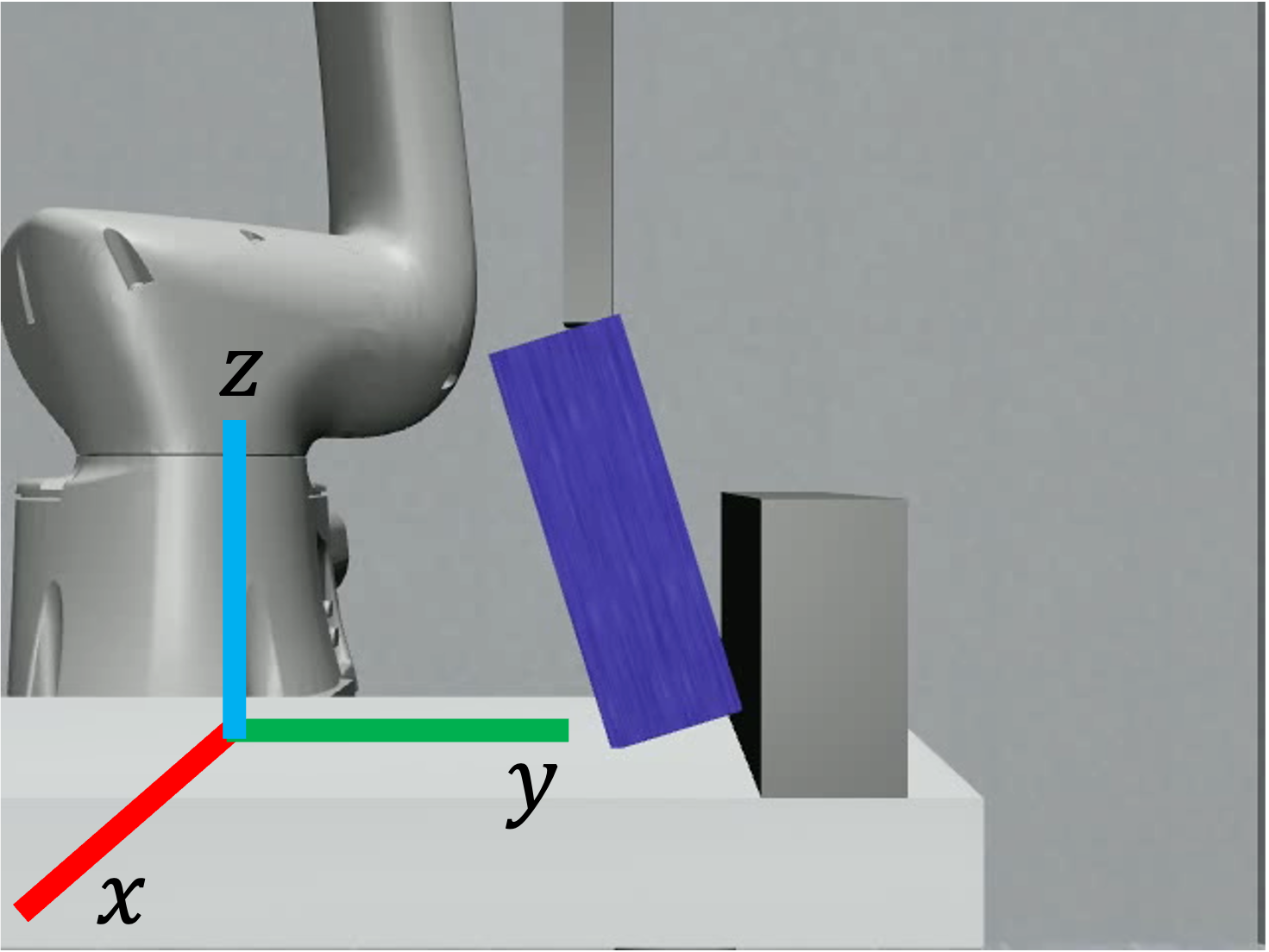}
    \caption{Definition of world frame used in this work. 
    }
\label{fig:frame}
\end{figure}

The coordinate is illustrated in \fig{fig:frame}. In this work, we operate within the SE(2) group, restricting manipulation to the $y-z$ plane. 

\subsection{Domain Randomization}
During the training of the teacher policy, we perform domain randomization and add sensor noises to robustify the policy, which is summarized in \tab{tab:dr_teacher_parameter}. 
\begin{table}[h]
    \centering
    \caption{Dynamics randomization and sensor noise. $\mathcal{N}(\mu, \sigma)$ denotes a Gaussian distribution with mean $\mu$ and standard deviation $\sigma$, and $\mathcal{U}(a, b)$ denotes a uniform distribution over the interval of $[a,b]$. A $+$ symbol indicates that the sampled noise is added to the original parameter value.}
    \begin{tabular}{cc} \toprule
           Parameter & Range \\ \midrule
        object mass &  $\mathcal{U}(0.04, 0.4)$ kg  \\
         friction for table and wall   &  $\mathcal{U}(0.01, 0.4)$  \\
         friction for objects   &  $\mathcal{U}(0.2, 0.7)$  \\
         friction for robots   &  $\mathcal{U}(0.7, 1.7)$  \\
         object size scale  &   $\mathcal{U}(0.95, 1.05)$ \\
         proportional gain $k_p$ in OSC  &  $\mathcal{U}(2000, 8000)$  \\
         derivative gain $k_d$ in OSC  & see below     \\ 
         initial object position along $y$-axis & +$\mathcal{U}(-0.015, 0.015)$ m \\ 
         initial robot position & +$\mathcal{U}(-0.015, 0.015)$ m \\ 
         object position observation noise & +$\mathcal{N}(0, 0.015)$  \\ 
         robot position observation noise  & +$\mathcal{N}(0, 0.00075)$  \\ 
         contact force observation noise & +$\mathcal{N}(0, 0.2)$  \\ 
         \bottomrule
    \end{tabular}
    \label{tab:dr_teacher_parameter}
\end{table}

For the derivative gain $k_d$ in operational space control (OSC) \cite{1087068}, we compute it based on the sampled proportional gain $k_p$ to achieve critical damping using the relation $k_d=2\sqrt{k_p}$.

It is worth noting that we represent object orientation using quaternions and apply domain randomization to account for sensor noise in orientation estimates.
Specifically, we perturb the ground-truth quaternion $\mathbf{q} \in \mathbb{R}^4$ by composing it with a small random rotation:
\[
\tilde{\mathbf{q}} = \delta \mathbf{q} \otimes \mathbf{q}
\]
where $\tilde{\mathbf{q}}$ is the noisy quaternion, $\delta \mathbf{q}$ is a perturbation quaternion, and $\otimes$ denotes quaternion multiplication.
The perturbation quaternion $\delta \mathbf{q}$ is constructed using a random axis-angle rotation. We first sample a unit axis $\mathbf{u} \in \mathbb{R}^3$ from a Gaussian distribution and normalize it:
\[
\mathbf{u} \sim \mathcal{N}(\mathbf{0}, \sigma_{\text{axis}}^2 \mathbf{I}), \quad \mathbf{u} \leftarrow \frac{\mathbf{u}}{\|\mathbf{u}\|}
\]
Next, we sample a rotation angle $\theta$ (in degrees) from a clipped Gaussian distribution:
\[
\theta \sim \text{clip}\left(\mathcal{N}(\mu_\theta, \sigma_\theta^2), -\theta_{\max}, \theta_{\max}\right)
\]
We then convert the axis-angle representation to a unit quaternion via the exponential map:
\[
\delta \mathbf{q} = \text{exp}(\theta \cdot \mathbf{u})
\]
In our implementation, we use the following parameters:
\[
\sigma_{\text{axis}} = 0.1, \quad \mu_\theta = 0^\circ, \quad \sigma_\theta = 2^\circ, \quad \theta_{\max} = 5^\circ
\]
This procedure injects bounded rotational noise into the observed quaternion while preserving unit norm and avoiding discontinuities.





\subsection{Termination Conditions}
An episode is terminated when any of the following conditions are met: 
\begin{enumerate}
    \item \textit{Successful task completion}: A trial is considered successful if the final orientation error satisfies $|\theta_e| \leq 0.087$ radians (i.e., \SI{5}{\degree}).
    \item \textit{Significant deviation from the SE(2) plane}: If the object's $x$-position $p_x$ deviates by more than \SI{0.05}{\meter} from its initial value, i.e., $|p_x - p_x(t=0)|\geq \SI{0.05}{\meter}$, or if the $z$-position drops below the table surface, $p_z \leq p_z^\text{table}$, the episode is terminated and a penalty of -100 is applied. 
    \item \textit{Timeout}: The episode exceeds the maximum number of steps as defined in \tab{tab:teacher-student-parameter}.
\end{enumerate}

\section{Student Estimator Details}
In this section, we provide details about the training procedure for the student estimator.

\subsection{Data Collection}\label{sec_appendix_data_collection_student}
To construct the dataset for student estimator training, we rollout the trained teacher policy in simulation and record the ground-truth privileged information, sensor observations, and corresponding object segmentation masks under domain randomization. 
We use the same range of domain randomization used during teacher policy training \tab{tab:dr_teacher_parameter}.
Since segmentation masks are not used during teacher policy training, we introduce additional uncertainties to simulate realistic conditions, including:
\begin{itemize}
    \item \textbf{Erosion/Dilation:} Morphological operations applied with random kernel sizes to simulate over- and under-segmentations.
    \item \textbf{Partial Mask Dropout:} Circular regions within the mask are randomly removed to mimic occlusions or partial detection failures.
    \item \textbf{Full Mask Dropout:} With a small probability, the entire mask is dropped (set to all zeros) to simulate complete sensor failure or occlusion.
    \item \textbf{Flip Noise:} Individual pixels are randomly flipped to simulate salt-and-pepper noise or detector flickering.
    \item \textbf{Edge Perturbation:} Object boundaries are randomly jittered to simulate segmentation boundary inaccuracies.
        \item \textbf{Spatial Augmentation (Affine):} Random affine transformations are applied to the mask, simulating viewpoint shifts and calibration noise.
    \item \textbf{Gaussian Blur:} A blur filter is applied to soften sharp edges and simulate optical imperfections.
\end{itemize}
The configuration of the segmentation domain randomization is summarized in \tab{tab:seg_noise}.
\begin{table}[h]
    \centering
    \caption{Segmentation mask domain randomization parameters used during student data collection.}
    \begin{tabular}{lcc} \toprule
        \textbf{Noise Type} & \textbf{Parameter} & \textbf{Value} \\ \midrule
        Erosion/Dilation & Probability for erosion/dilation & 0.7 \\
        & Kernel size choices & $\{3, 5, 7\}$ \\
        & Erosion vs. dilation split & 0.5 \\
        Random Holes & Number of holes & 3 \\
        & Hole radius range & $[3, 9]$ pixels \\
        & Hole probability & 0.5 \\
        Full Mask Dropout & Probability & 0.05 \\
        Flip Noise & Pixel flip probability & 0.01 \\
        Edge Perturbation & Edge noise probability & 0.75 \\
        & Edge point noise probability & 0.1 \\
        Spatial Augmentation (Affine)
        & Rotation range & $\pm 2.5^\circ$ \\
        & Translation range & $\pm 7.5\%$ \\
        & Scaling range & $[0.95,\ 1.05]$ \\
        \bottomrule
    \end{tabular}
    \label{tab:seg_noise}
\end{table}

\subsection{Student estimator training}

Given the dataset collected in Section \ref{sec_appendix_data_collection_student}, we train a student estimator composed of a CNN followed by a TCN. 
The CNN takes as input a binary segmentation mask of size $1 \times 480 \times 640$ and consists of three convolutional layers with kernel sizes of $(3, 3, 3)$, and strides of $(2, 2, 1)$, and output channels of $(16, 32, 64)$, respectively. 
An adaptive average pooling layer reduces the spatial dimensions to $8 \times 8$, followed by a fully connected layer that produces a $1 \times 128$ feature vector. 
The TCN processes the temporal sequence of CNN features concatenated with proprioceptive and force features. It consists of three layers of 1D dilated causal convolutions, each with 128 channels and a kernel size of 2, and dilation rates of 1, 2, and 4.
We consider two types of privileged information: time-invariant dynamics parameters (i.e., mass and size of the object), and time-varying values such as the object pose. 
To accommodate this distinction, the student estimator employs two separate fully connected layers---one for predicting the time-invariant variables and another for the time-varying privileged quantities (e.g., object pose). The output dimensions of each head match the corresponding target variables. We find that this separation leads to improved estimation performance. 

\begin{figure}[t]
    \centering
    \includegraphics[width=0.5\textwidth]{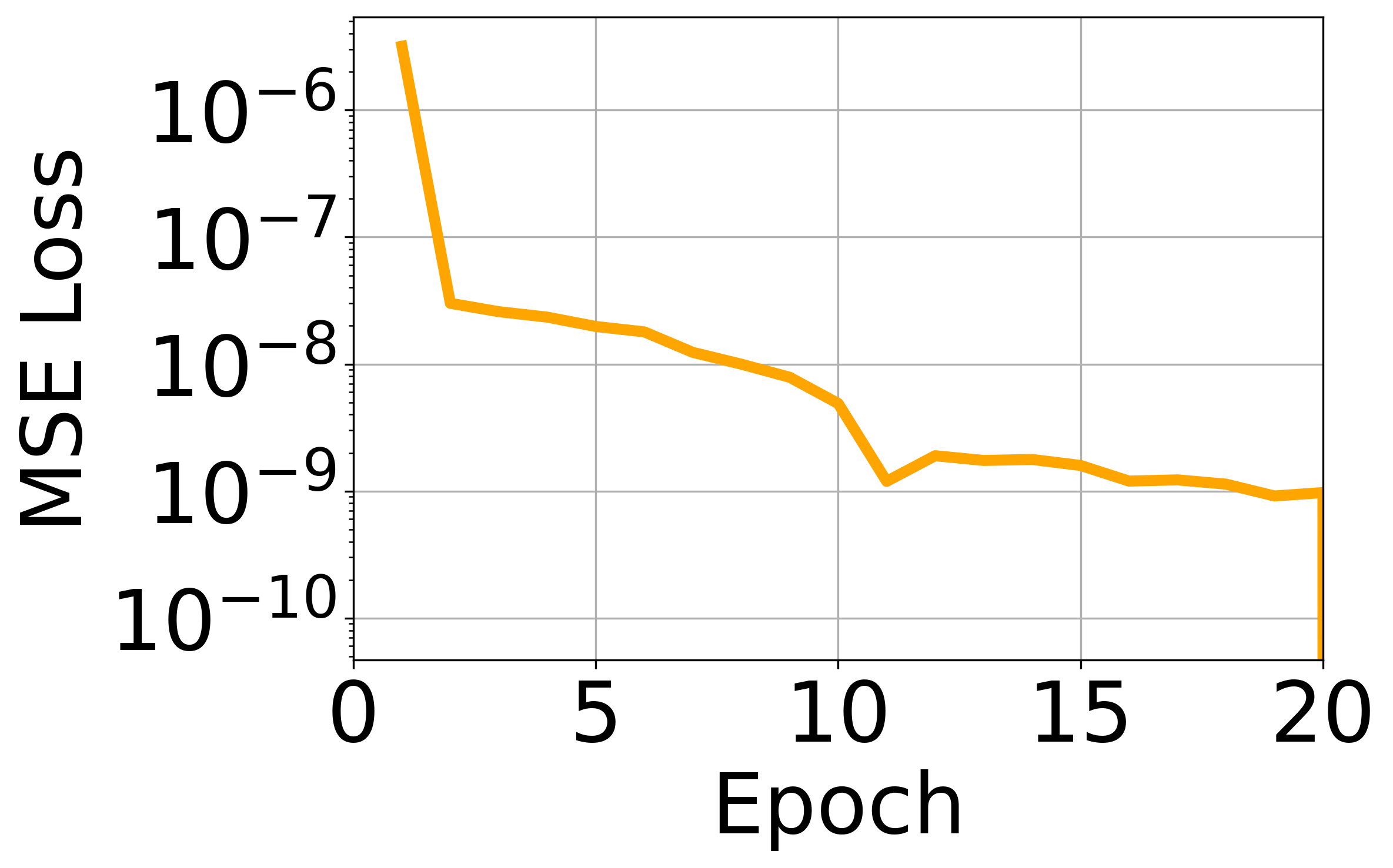}
    \caption{Student estimator validation loss over epochs. 
    }
\label{fig:studet_training_data}
\end{figure}

Then, the model is trained by minimizing the mean square error between the ground-truth and predicted values by the student estimator. \fig{fig:studet_training_data} shows the learning curve of the validation loss during training. 
The hyperparameters used for training the student estimator are summarized in \tab{tab:student-parameter}.
\begin{table}[h]
    \centering
    \caption{Hyperparameter setup for student estimator.}
    \begin{tabular}{cc} \toprule
           Parameter & Value \\ \midrule
        total \# of epochs &  20  \\
        batch size &  256  \\
        initial learning rate &  1e-3  \\
         learning rate schedule &  ReduceLROnPlateau from PyTorch \\
         optimizer & Adam  \\
         \bottomrule
    \end{tabular}
    \label{tab:student-parameter}
\end{table}

\section{Ablation Study}

\subsection{Effect of linear and quadratic reward terms during teacher policy training}
In Section \ref{sec:method}, we mention that using linear and quadratic terms in $r_p$ in \eq{reward_eq} is important to ensure that the robot completes the pivoting task. 
To validate this claim, we conducted an ablation study using dynamics-conditioned RL, evaluating three reward variants: (1) linear only, (2) quadratic only, and (3) both linear and quadratic terms in \( r_p \), under settings with and without domain randomization.  \tab{tab:linear_vs_quadratic_reward} shows the mean and standard deviation of the terminal object angle over 50 evaluation episodes.

When domain randomization is disabled, the policy trained with the linear term alone in \( r_p \) successfully completes the pivoting task. In contrast, using only the quadratic term leads to task failure, likely due to the difficulty in reward shaping—quadratic rewards are sparse and less informative during early training.
On the other hand, when domain randomization is enabled, 
policies trained with only the linear term exhibit significantly degraded performance. In this case, combining linear and quadratic terms improves performance substantially. We hypothesize that the quadratic component offers a stronger gradient signal when the agent is close to the goal, helping to overcome the increased noise due to domain randomization.


\begin{table}[h]
    \centering
    \caption{Comparison of terminal object angle using different reward formulation with/without domain randomization. In the terminal angle, we show its mean with standard deviation over 50 episodes. }
    \begin{tabular}{rcr} \toprule
           Reward type & Enable domain randomization & Terminal angle [deg] \\ \midrule
        Linear term & No & 88.1 $\pm 0.21$   \\
         Quadratic term & No & 0.0 $\pm 0.10$   \\
         Linear + Quadratic term & No & 88.9 $\pm 0.20$   \\ \midrule
                 Linear term & Yes & 70.1 $\pm 0.59$   \\
         Quadratic term & Yes & 0.0 $\pm 0.71$   \\
         Linear + Quadratic term & Yes & 88.2 $\pm 0.44$   \\
         \bottomrule
    \end{tabular}
    \label{tab:linear_vs_quadratic_reward}
\end{table}

\subsection{Pivoting with wall task without domain randomization}

\begin{figure}[h]
    \centering
    \includegraphics[width=0.5\textwidth]{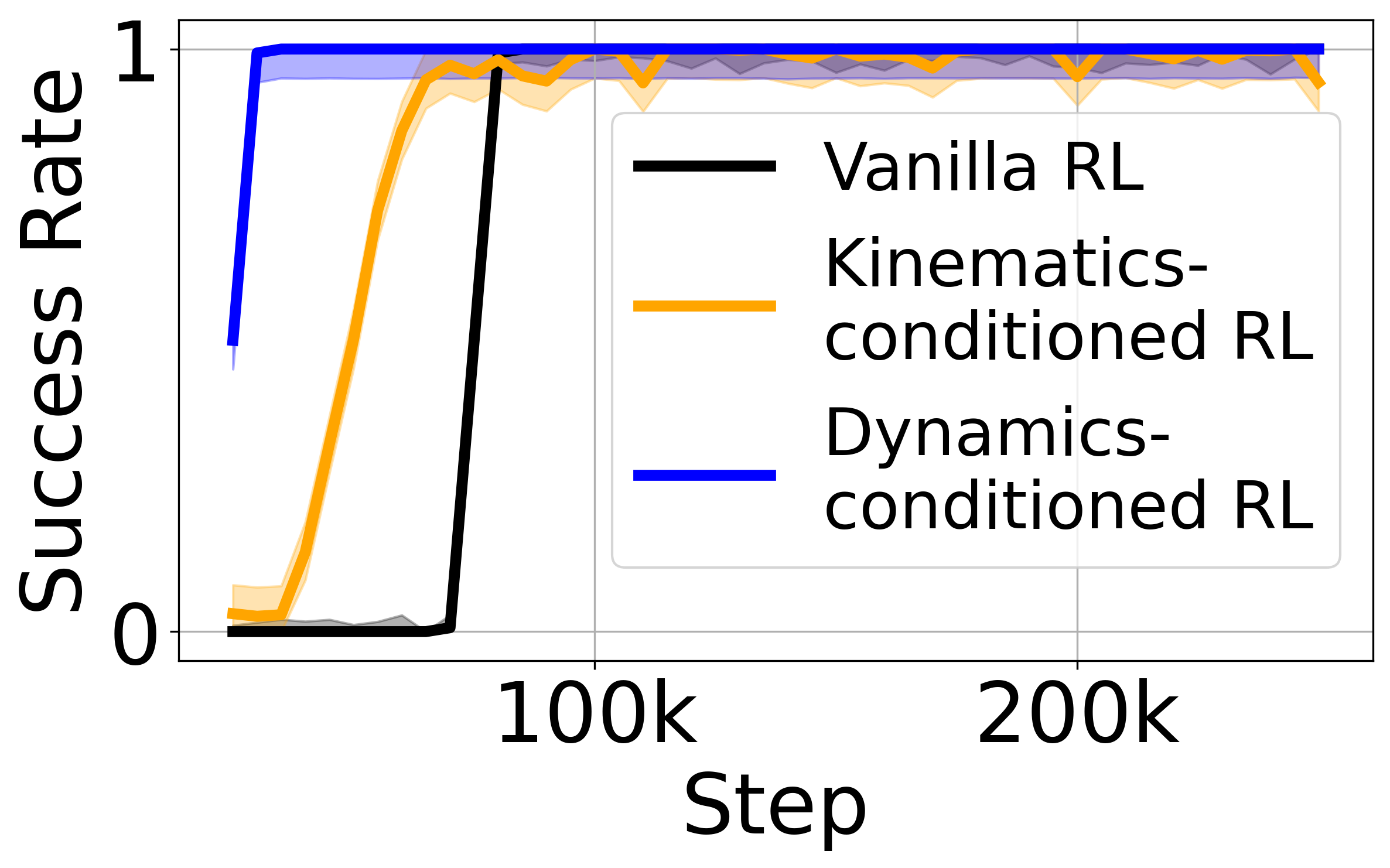}
    \caption{Learning curves for different RL training runs for pivoting-with-wall task. Solid lines indicate average success rates, and shaded regions denote standard deviation across three different random seeds. Every $10$k step, the current policy is evaluated over $50$ episodes, and the success rate is plotted.  
    }
\label{fig:pivoting_no_dr_traing}
\end{figure} 

In Section \ref{sec:result}, we present the result of the training curve using different RL training runs for two tasks. For the results in \fig{sample_eff}, we consider domain randomization, and thus it is possible that the pivoting with external wall task could not be trained due to the large domain randomization. Hence, we show the result for the pivoting with wall task under no domain randomization as shown in \fig{fig:pivoting_no_dr_traing}.

\fig{fig:pivoting_no_dr_traing} shows that all RL using different reward equations could successfully learn the skill. 
Among them, dynamics-conditioned RL exhibits the fastest learning rate. This confirms that while vanilla RL can succeed when the training environment is noise-free, providing dynamics-consistent demonstrations significantly improves the learning efficiency by offering more informative reward signals.


We emphasize that for the pivoting without wall task, even under no domain randomization, vanilla RL and kinematically-conditioned RL fail to learn. This supports our claim that non-prehensile manipulation tasks have very narrow feasible action regions. Therefore, leveraging demonstrations that satisfy complex contact constraints plays an important role in improving learning efficiency.

\subsection{Student Estimator Performance}

\begin{figure}[h]
    \centering
    \begin{subfigure}{0.32\textwidth}
        \centering
        \includegraphics[width=1.0\linewidth]{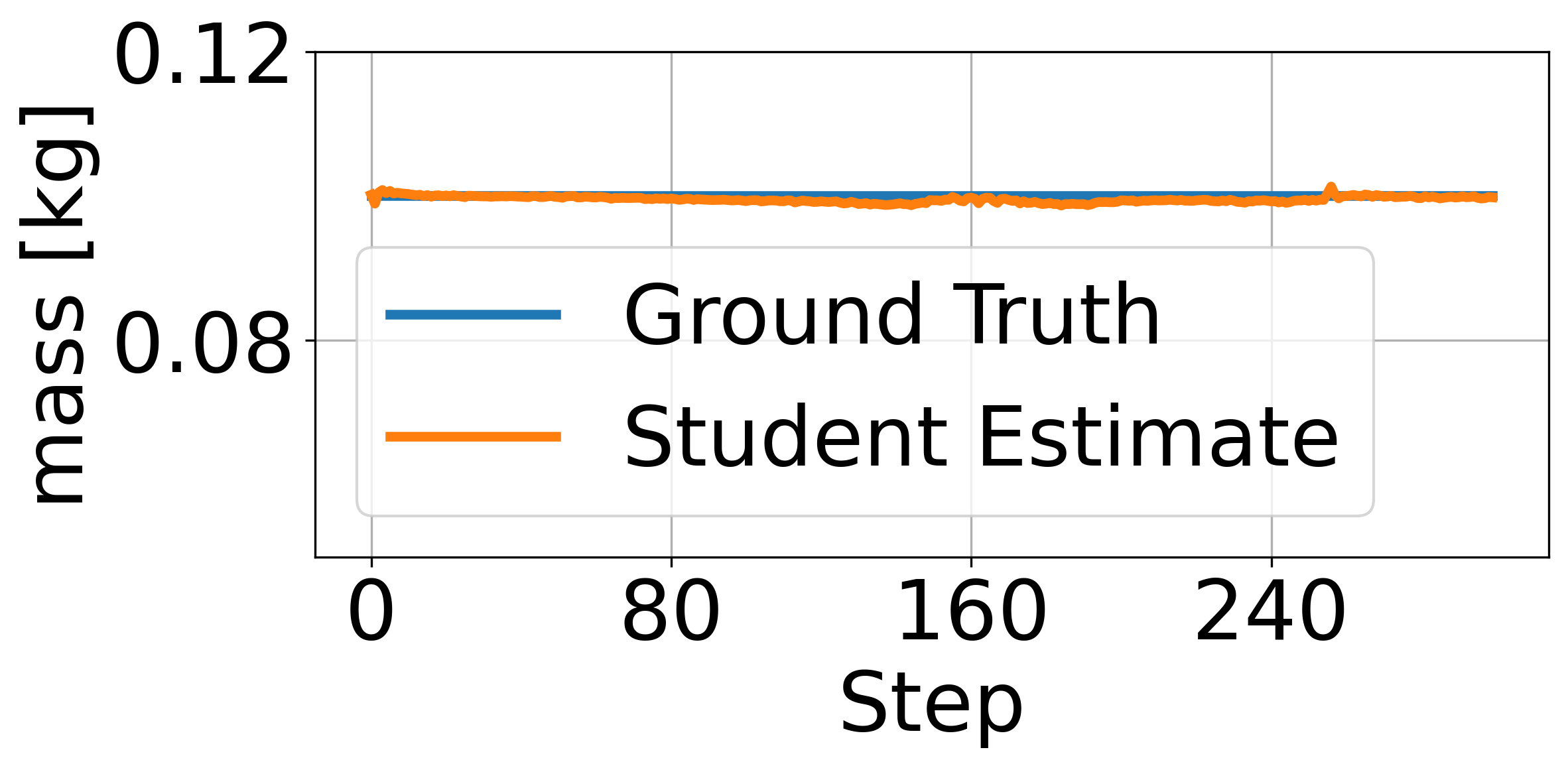}
    \end{subfigure}
    \begin{subfigure}{0.32\textwidth}
        \centering
        \includegraphics[width=1.0\linewidth]{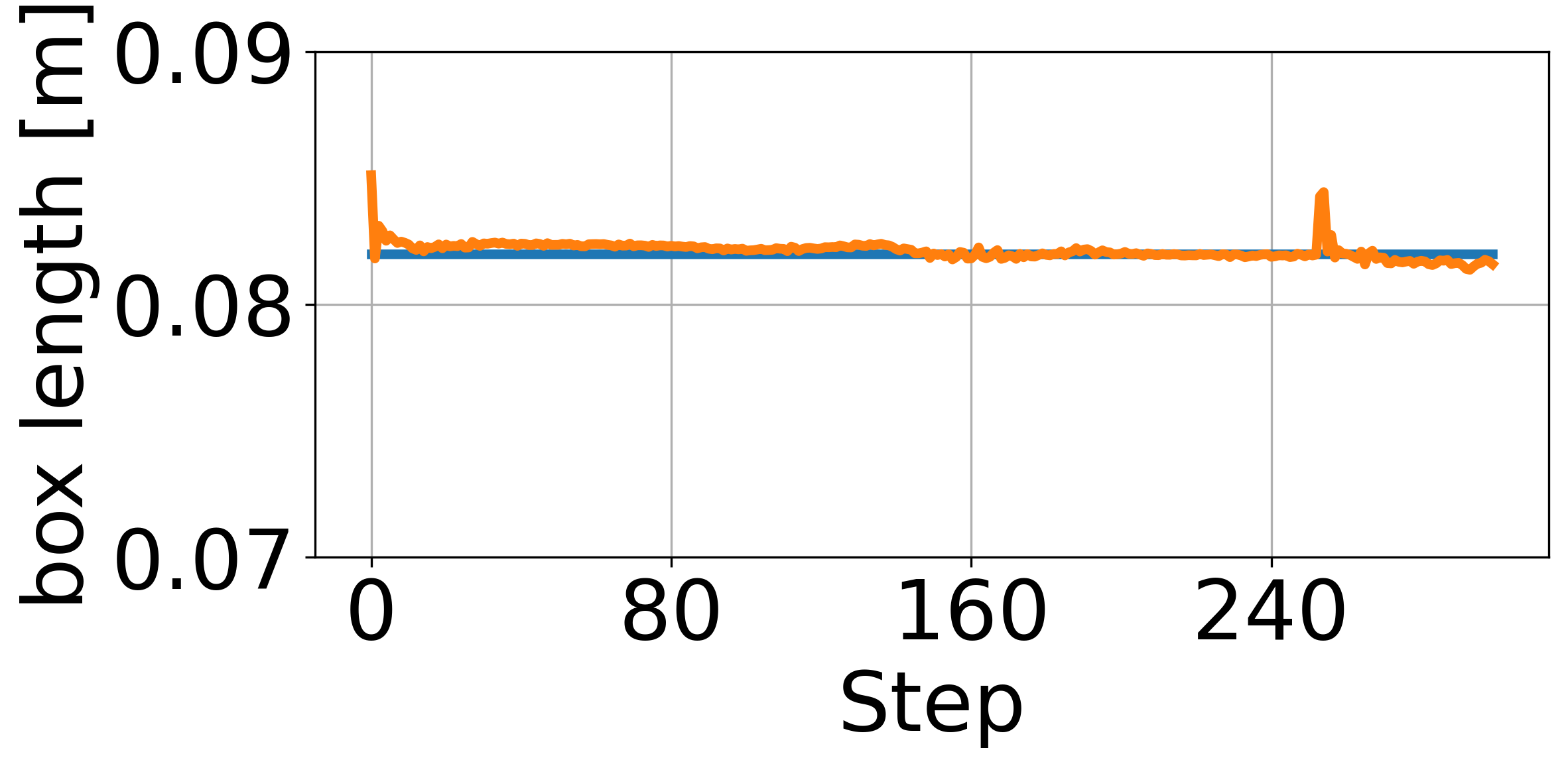}
    \end{subfigure}
        \begin{subfigure}{0.32\textwidth}
        \centering
        \includegraphics[width=1.0\linewidth]{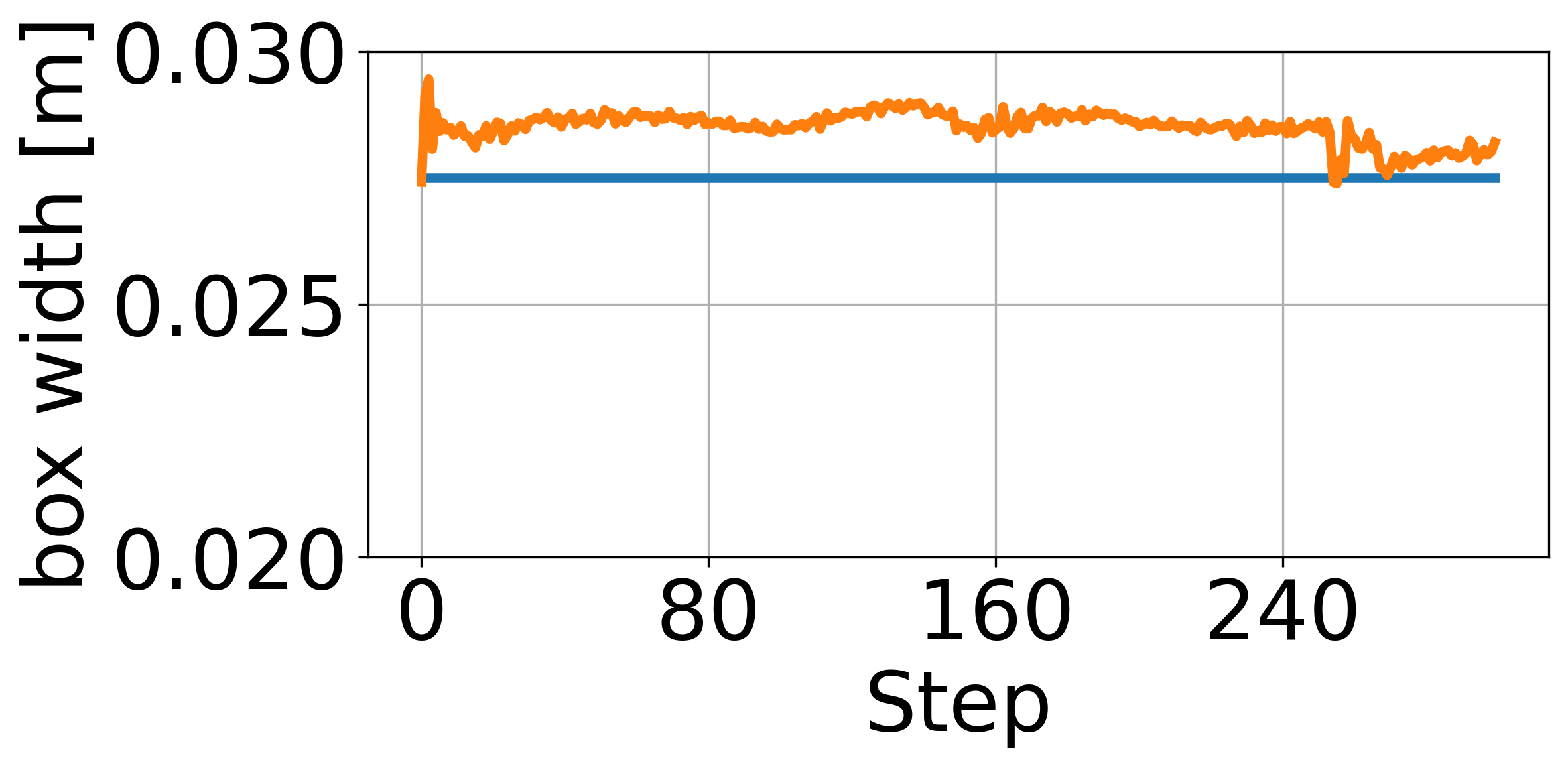}
    \end{subfigure}
            \begin{subfigure}{0.32\textwidth}
        \centering
        \includegraphics[width=1.0\linewidth]{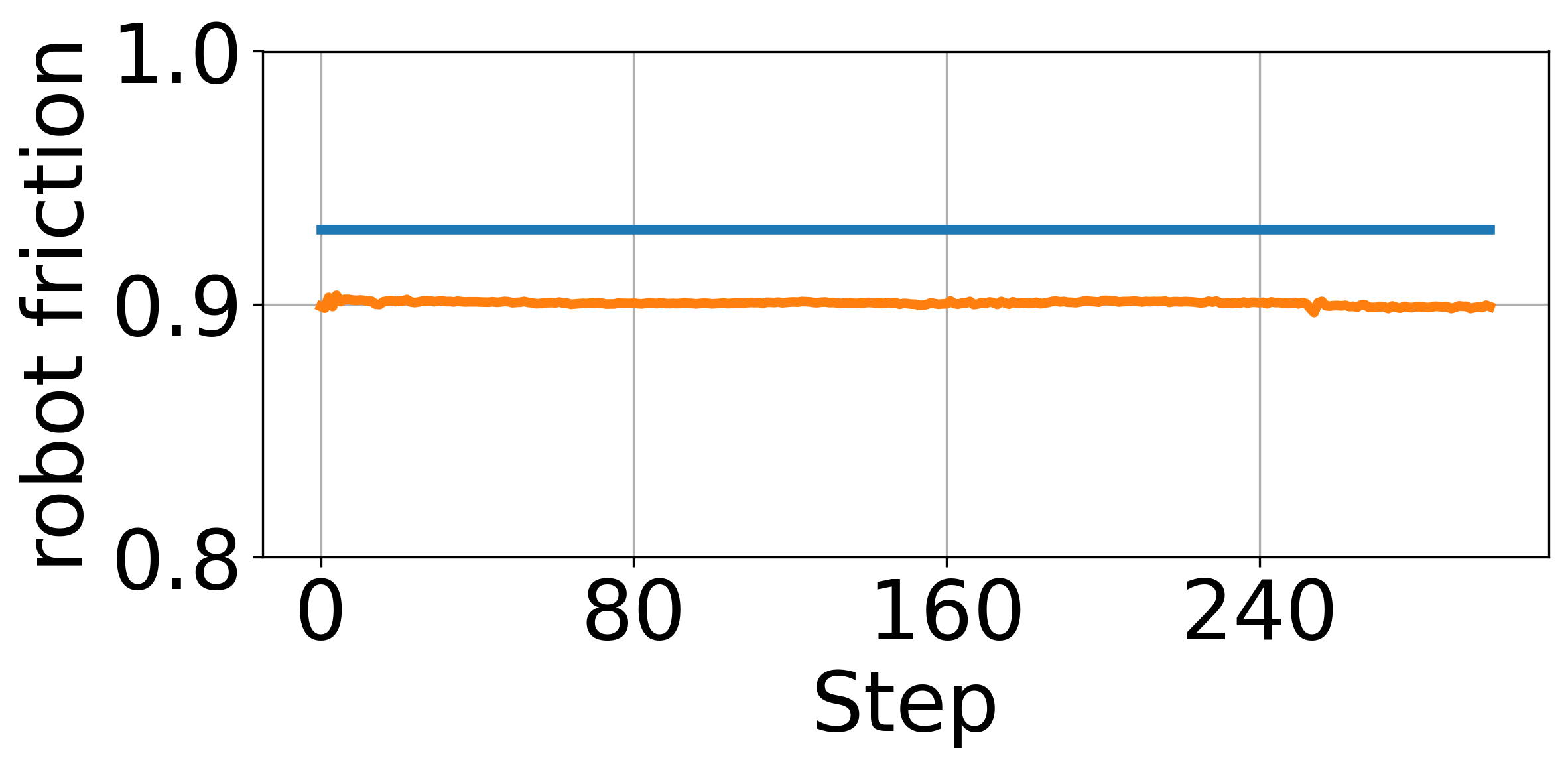}
    \end{subfigure}
                    \begin{subfigure}{0.32\textwidth}
        \centering
        \includegraphics[width=1.0\linewidth]{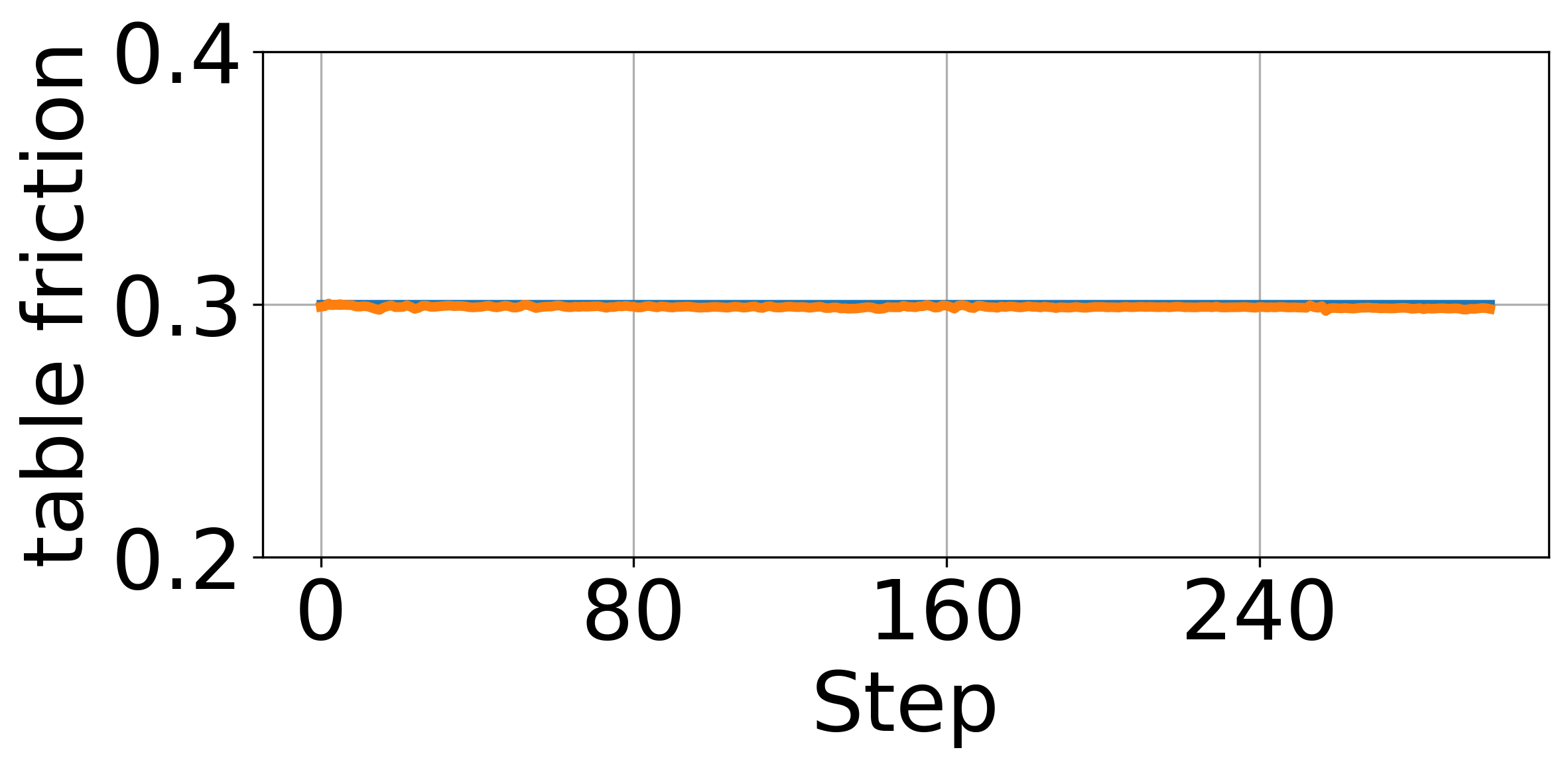}
    \end{subfigure}
    \caption{Comparison of our student estimator's predictions and the ground truth for the box mass, the box length, the box width, the robot friction constant, and the table friction constant, for the pivoting with a wall.}
    \label{student-est-fig_all}
\end{figure}

In Section \ref{sec:result}, we present a subset of our student estimator results due to page limitations. We show the remaining privileged information figures in \fig{student-est-fig_all}. 
Overall, we observe that our student estimator successfully predicts the privileged information with reasonable accuracy. 



\subsection{Sim-to-Real Transfer}

\begin{figure}[h]
    \centering
    \begin{subfigure}{0.495\textwidth}
        \centering
        \includegraphics[width=1.0\linewidth]{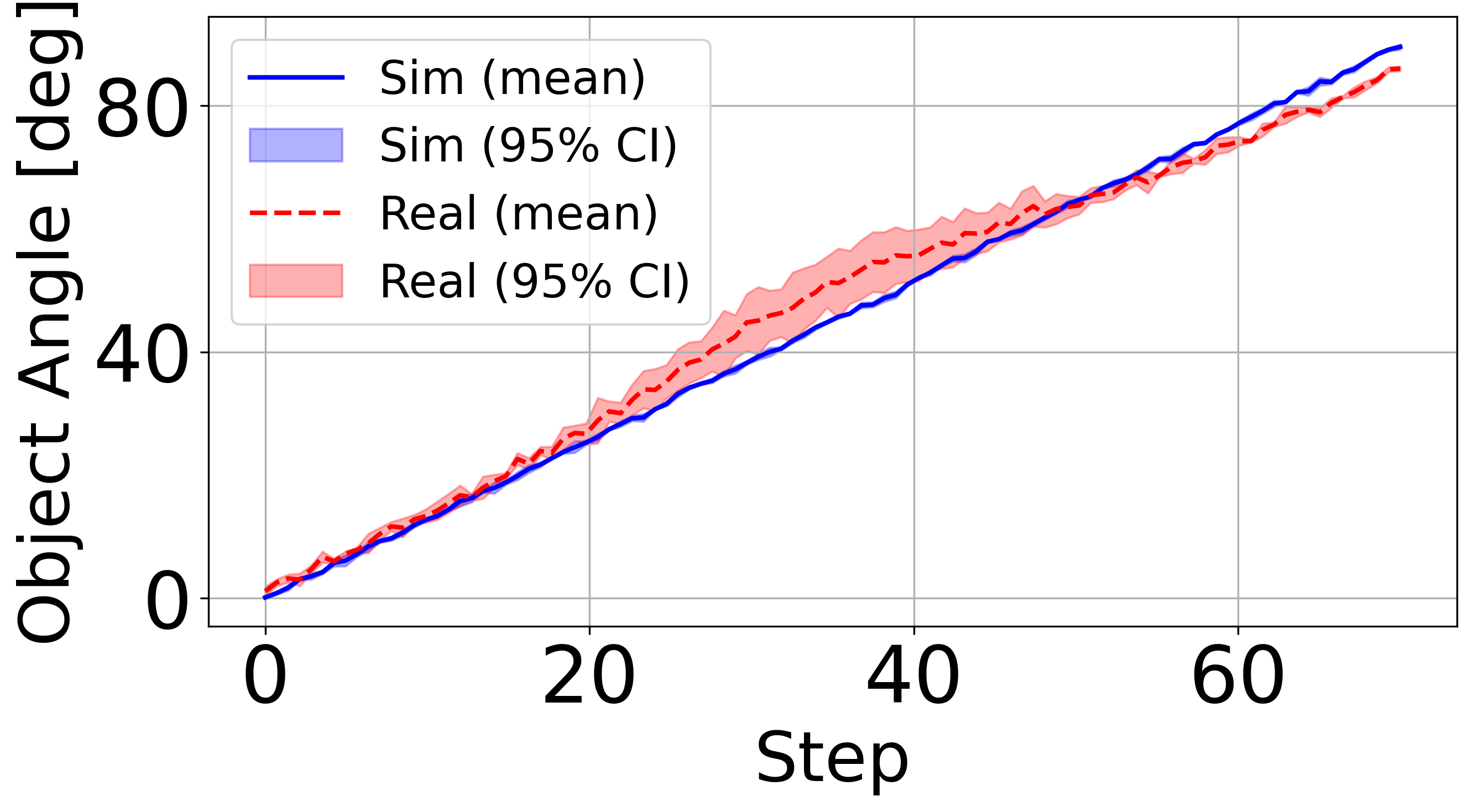}
        \caption{Pivoting with external wall}
    \end{subfigure}
    \begin{subfigure}{0.495\textwidth}
        \centering
        \includegraphics[width=1.0\linewidth]{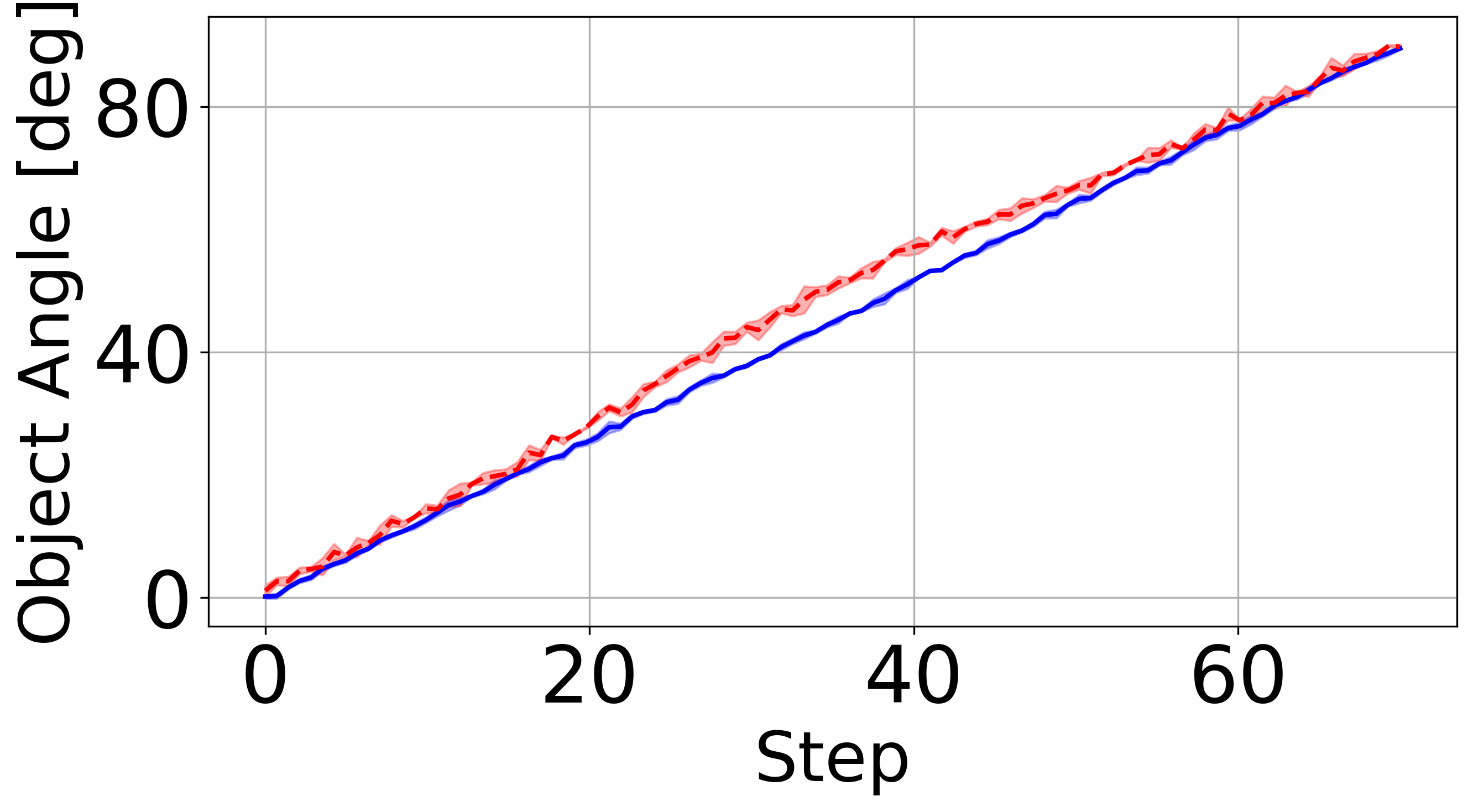}
\caption{Pivoting without external wall}
    \end{subfigure}
        \hfill
    \caption{
    Comparison of the object angle in simulation and the real-world during pivoting. We execute the same policy both in simulation and in hardware and collect the object orientation during manipulation over 3 trials.
    Due to sensor discrepancies and physical modeling differences (i.e., sim-to-real gap), the resulting actions and motion can differ between simulation and hardware.
    }
    \label{sim2real}
\end{figure}

To evaluate sim-to-real transfer, we deploy the learned dynamics-conditioned RL policy on both the simulation and physical hardware for the two pivoting tasks. The resulting object orientation trajectories over three trials are shown in \fig{sim2real}.


Overall, although there is some sim-to-real gap for both tasks, the robot could successfully perform the tasks on the physical hardware as shown in the attached supplemental video. 
We observe a larger sim-to-real gap for the pivoting with external wall task than the pivoting without wall task. 
This is because for the pivoting with wall task, the object induces the sliding contact between the object and the wall, and between the object and the table, which are relatively challenging to model precisely in simulator (e.g., MuJoCo), leading to a larger sim-to-real gap. 
In contrast, the pivoting-without-wall task does not involve sliding contacts, resulting in better sim-to-real transfer.
\end{document}